\documentclass[runningheads]{llncs}

\usepackage{eccv}

\usepackage{eccvabbrv}

\usepackage{graphicx}
\usepackage{booktabs}
\usepackage{multirow}
\usepackage{wrapfig}
\usepackage[accsupp]{axessibility}  

\usepackage[breaklinks,colorlinks,citecolor=eccvblue]{hyperref}
\usepackage{orcidlink}

\newcommand\blfootnote[1]{%
  \begingroup
  \renewcommand\thefootnote{}\footnote{#1}%
  \addtocounter{footnote}{-1}%
  \endgroup
}

\definecolor{todo}{RGB}{166, 45, 45}
\definecolor{mygrey}{RGB}{220,220,220}
\usepackage{colortbl}
\usepackage{algorithm}
\usepackage{listings}
\usepackage{algpseudocode}
\usepackage{graphicx}
\usepackage{booktabs}
\usepackage{multicol}
\usepackage{bbding} 
\usepackage{etoolbox}
\usepackage{listings}

\newcommand\algcomment[1]{\def\@algcomment{\footnotesize#1}}

\begin{document}

\title{ConGeo: Robust Cross-view Geo-localization across Ground View Variations}


\author{Li Mi\inst{1*}\orcidlink{0000-0002-4886-2430} \and
Chang Xu\inst{2,1*\text{†}}\orcidlink{0000-0002-3078-0496} \and
Javiera Castillo-Navarro\inst{1}\orcidlink{0000-0003-4917-5103} \and
Syrielle Montariol\inst{1}\orcidlink{0000-0003-1355-8778} \and \\
Wen Yang\inst{2}\orcidlink{0000-0002-3263-8768} \and
Antoine Bosselut\inst{1}\orcidlink{0000-0001-8968-9649} \and
Devis Tuia\inst{1}\orcidlink{0000-0003-0374-2459}
}

\authorrunning{L.~Mi and C.~Xu et al.}
\institute{EPFL\email{} 
\and Wuhan University\
\email{}\\
\url{https://eceo-epfl.github.io/ConGeo/}
}

\maketitle
\begin{abstract}
\blfootnote{$^{*}$ Equal contribution \quad $^{\text{†}}$ Corresponding author (xuchangeis@whu.edu.cn)} Cross-view geo-localization aims at localizing a ground-level query image by matching it to its corresponding geo-referenced aerial view. In real-world scenarios, the task requires accommodating diverse ground images captured by users with varying orientations and reduced field of views (FoVs). 
However, existing learning pipelines are orientation-specific or FoV-specific, demanding separate model training for different ground view variations. 
Such models heavily depend on the North-aligned spatial correspondence and predefined FoVs in the training data, compromising their robustness across different settings.
To tackle this challenge, we propose \textbf{ConGeo}, a single- and cross-view \textbf{Con}trastive method for \textbf{Geo}-localization: it enhances robustness and consistency in feature representations to improve a model's invariance to orientation and its resilience to FoV variations, by enforcing proximity between ground view variations of the same location. As a generic learning objective for cross-view geo-localization, when integrated into state-of-the-art pipelines, ConGeo significantly boosts the performance of three base models on four geo-localization benchmarks for diverse ground view variations and outperforms competing methods that train separate models for each ground view variation.
\end{abstract}

\section{Introduction}
\label{sec:intro}
Given an image captured at ground level, cross-view geo-localization (CVGL) aims to determine the location of the image by referring to its corresponding aerial view~\cite{crossview_cvpr_2013,jointloc_2020_cvpr,safa_nips_2019,transgeo_cvpr_2022,sample4geo_2023_iccv}. As an important auxiliary positioning technique, the task enables noisy GPS correction~\cite{zamir2010accurate} and offers practical applications in fields such as navigation~\cite{li2019cross} and autonomous driving~\cite{fervers2023uncertainty}.

\begin{figure}[t]
    \centering
    \includegraphics[width=0.98\linewidth]{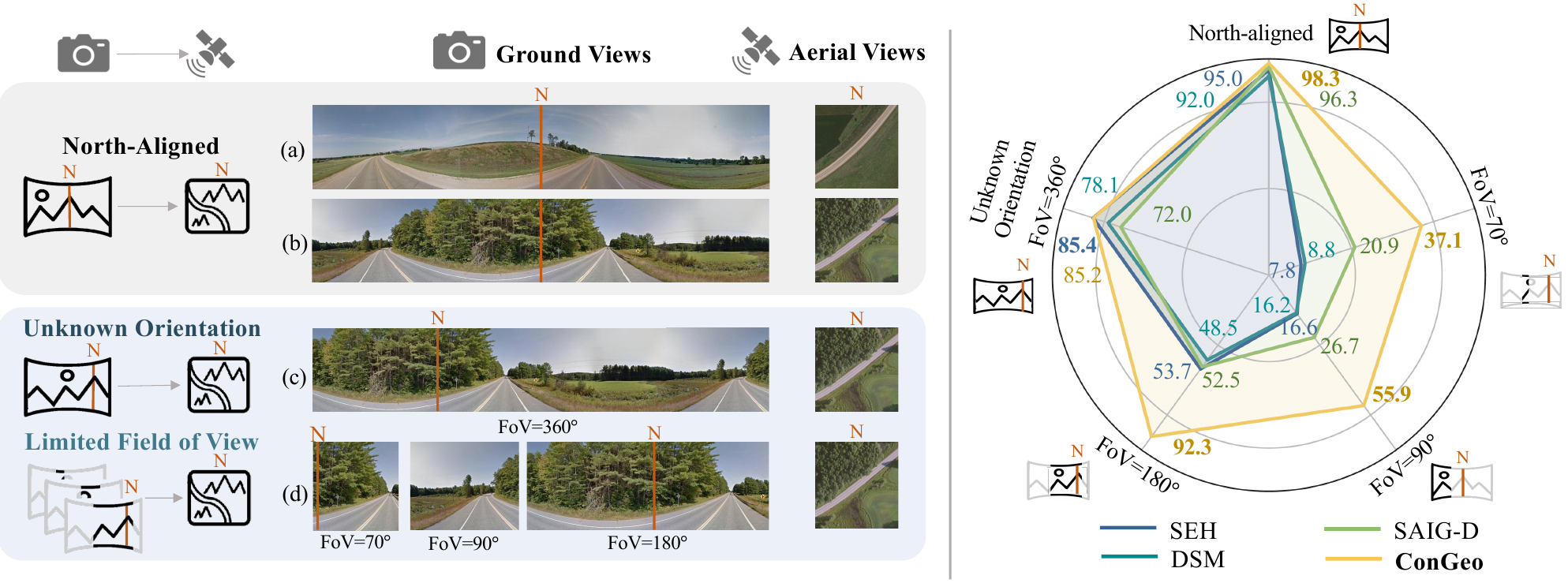}
    \caption{\textbf{ConGeo boosts the robustness across ground view variations}: North-aligned, unknown orientation (FoV=360$^{\circ}$) and limited field of views (FoV=70$^{\circ}$, 90$^{\circ}$, and 180$^{\circ}$). We compare with SEH~\cite{guo2022softSEH}, DSM~\cite{jointloc_2020_cvpr} and SAIG-D~\cite{zhu2023simplesaigd} and report Top-1 Recall on the CVUSA~\cite{cvusa_cvpr_2015} dataset, one of the geo-localization benchmarks.} 
    \label{fig:intro}
\end{figure}

Cross-view geo-localization is often addressed as a retrieval task, where a ground-level image acts as the query and a geo-tagged overhead image as the reference. In real-world applications, it often requires a high generalization ability, to handle diverse ground view image variations, including varying orientations and reduced field of view (FoV). In the past, models were trained and evaluated under idealised settings, where both views are aligned to the North~\cite{cvusa_cvpr_2015, lending_2019_CVPR, vigor_cvpr_2021} which only covers limited scenarios in the real-world challenges (see \textbf{North-aligned} setting in the left panel of Fig.~\ref{fig:intro}, rows (a) and (b)). Recent works~\cite{jointloc_2020_cvpr, zhu2023simplesaigd} extend existing North-aligned datasets to more challenging settings, cyclically shifting a ground view panorama by a random angle to obtain orientation variations, or further reducing the FoV from 360$^{\circ}$ to 70$^{\circ}$, 90$^{\circ}$, or 180$^{\circ}$ to simulate limited ground view information (see left panel of Fig.~\ref{fig:intro}, \textbf{Unknown Orientation} setting in rows (c) and \textbf{Limited FoV} setting in row (d)). However, they train and evaluate models separately for each setting. This limits the generalization ability of the model in real-world applications, where the FoV and orientation of the query image are often unknown. Moreover, the orientation-specific or FoV-specific training prevents the model from representing robust features across different settings. Instead, the models are often biased toward training data to capture specific spatial correspondence that is not generalizable across ground view variations. For example, in the left panel of Fig.~\ref{fig:intro} (rows (a) and (b)), the two aerial views share similar road directions, which can also be found in the corresponding ground views. Our analysis shows such North-aligned spatial correspondence in the training data can serve as a shortcut for the model to improve the specific setting but may sacrifice the orientation invariance and feature consistency towards ground view variations.

To address this challenge, we introduce \textbf{ConGeo}, a model-agnostic \textbf{Con}-trastive learning pipeline for cross-view \textbf{Geo}-localization. 
ConGeo follows the intuition that a robust model should capture consistent features for the same location regardless of orientations or FoVs and identify the same reference image across ground view variations. To achieve that, we design two contrastive objectives to enforce the proximity between ground view variations and their original representation: a single-view contrastive objective and a cross-view one.
The ground view contrastive loss works to align ground view variants with the original images (North-aligned, full panorama), and the aerial view contrastive loss minimizes the representation disparity between the aerial image and its augmented counterpart. 
The cross-view (ground view and aerial view) contrastive loss further aligns the query image variants with reference aerial images. By disrupting the initial geometric correspondence and mitigating potential shortcuts found in the training data, ConGeo compels the model to focus on learning coherent features across different modalities and view variations.

We run extensive experiments on four geo-localization benchmarks, demonstrating that ConGeo outperforms existing methods across ground view variations. Moreover, our results highlight three insights: \textbf{(1)} ConGeo empowers a single model to handle various ground view variations, using a model-agnostic learning pipeline. ConGeo consistently outperforms comparison methods under orientation-specific or FoV-specific training by a large margin when facing ground view orientation and FoV shifts (The right panel of Fig.~\ref{fig:intro}, Table~\ref{tab: fov}). Meanwhile, ConGeo demonstrates strong versatility, as it can be plugged into different geo-localization pipelines (Table~\ref{tab:plugin}). \textbf{(2)} We demonstrate the advantages of the proposed contrastive learning pipeline over targeted data augmentation when facing ground view variations (Table~\ref{tab:aug}), especially unseen ones (Table~\ref{tab:unseen}). \textbf{(3)} Our analysis reveals that the trade-off between the model's focus on geometric or semantic cues for matching images from different views affects its robustness among variations (Fig.~\ref{fig:attention360} and~\ref{fig:attention}).

\section{Related Works}
\label{sec:related_works}
\subsection{Cross-view Geo-localization}

Cross-view geo-localization has been an active field of research over the last decade~\cite{crossview_cvpr_2013,transgeo_cvpr_2022,sample4geo_2023_iccv,safa_nips_2019}. We can distinguish two categories of works: first is 
\textbf{geo-localization with North alignment},  the classical evaluation setting for cross-view image geo-localization. The standard approach uses a siamese network~\cite{siamese_nips_1993} to encode the ground view and the aerial images and optimizes a well-designed loss to align the feature embeddings.
A series of works pointed out the importance of spatial correspondence in paired images and incorporated such kind of prior into the pipeline; SAFA~\cite{safa_nips_2019} designed a polar transformation that explicitly aligns two domains and used spatial attention to facilitate network learning; Liu et al.~\cite{lending_2019_CVPR} encoded the orientation correspondence between views as additional input for the network, allowing the model to be orientation-discriminative. 
Recent studies~\cite{zhu2023simplesaigd, transgeo_cvpr_2022, sample4geo_2023_iccv} use more advanced architectures to model global relationships between views, TransGeo~\cite{transgeo_cvpr_2022} uses learnable position encoding that implicitly learns the cross-view correspondence, Sample4Geo~\cite{sample4geo_2023_iccv} uses the ConvNeXt~\cite{convnext_cvpr_2022} as feature extractor and designs GPS-based and similarity-based samplers for hard negative sample mining.
\\Second is 
{\textbf{geo-localization with variations in orientation and FoV}: compared to strictly North-aligned panoramas, ground images with unknown orientations and limited FoV (\textit{e.g.}, captured by a smartphone or car camera) are more readily available, but make the task more challenging. Recent studies~\cite{revisiting_2021_WACV, jointloc_2020_cvpr, global_2022_wacv} gradually draw attention to this more realistic scenario. Specifically, DSM~\cite{jointloc_2020_cvpr} proposed to crop out the query-activated region in the reference image after orientation estimation for accurate matching. Rodrigues et al.~\cite{global_2022_wacv} customized a pipeline for the retrieval under limited FoVs, where the limited FoV images are cropped from the original image as a form of data augmentation for both views. Despite the steady progress, these methods still face limitations: First, specific FoV information is required for aerial image/feature cropping during both training and testing, while in real-world data this strong prior is unknown, especially during testing. Second, models trained on a specific setting cannot generalize and do not perform satisfactorily on all FoV scenarios. Thus several specialised models are trained for each setting.

Instead, we propose a method that does not rely on the exact FoV information during training nor testing, and only needs to be trained end-to-end once to achieve competitive results in all settings (see the right panel of Fig.~\ref{fig:intro}).
}

\subsection{Contrastive Learning for Geo-localization}
Contrastive learning has shown impressive performance for self-supervised and supervised learning in computer vision~\cite{chen2020simclr, he2020moco, khosla2020supcon,makesforcontrastive_nips_2020}, including image classification~\cite{contrastive_classification_2021_cvpr}, object detection~\cite{detco_iccv_2021}, and vision-language tasks~\cite{vqa_contrastive_emnlp_2020}. Existing methods use either cross-modal objectives to align different modalities (\textit{e.g.}, CLIP~\cite{radford2021learningclip}) or single-modal ones to enhance single-modal representation (\textit{e.g.}, SLIP~\cite{mu2022slip}). 
In CVGL, the triplet loss~\cite{hu2018cvmCVMNET, lending_2019_CVPR, optimaltransport_2019, jointloc_2020_cvpr, transgeo_cvpr_2022} has been a standard choice to solve the cross-view image retrieval problem in a contrastive way. More recently, 
InfoNCE~\cite{oord2018representation, sample4geo_2023_iccv, zhu2023simplesaigd} has shown to be an efficient objective for the task. However, contrastive learning in these works is limited to aligning feature embeddings between the vanilla ground view and aerial images to match image pairs; while it has shown a powerful ability for building view-invariant and robust representations~\cite{he2020moco,makesforcontrastive_nips_2020,chen2020simclr}. In this work, we further leverage contrastive learning by using both single- and cross-view contrastive objectives, achieving more robust feature representations shared by ground view variants.

\section{ConGeo}

\begin{figure*}[t]
    \centering
    \includegraphics[width=0.98\linewidth]{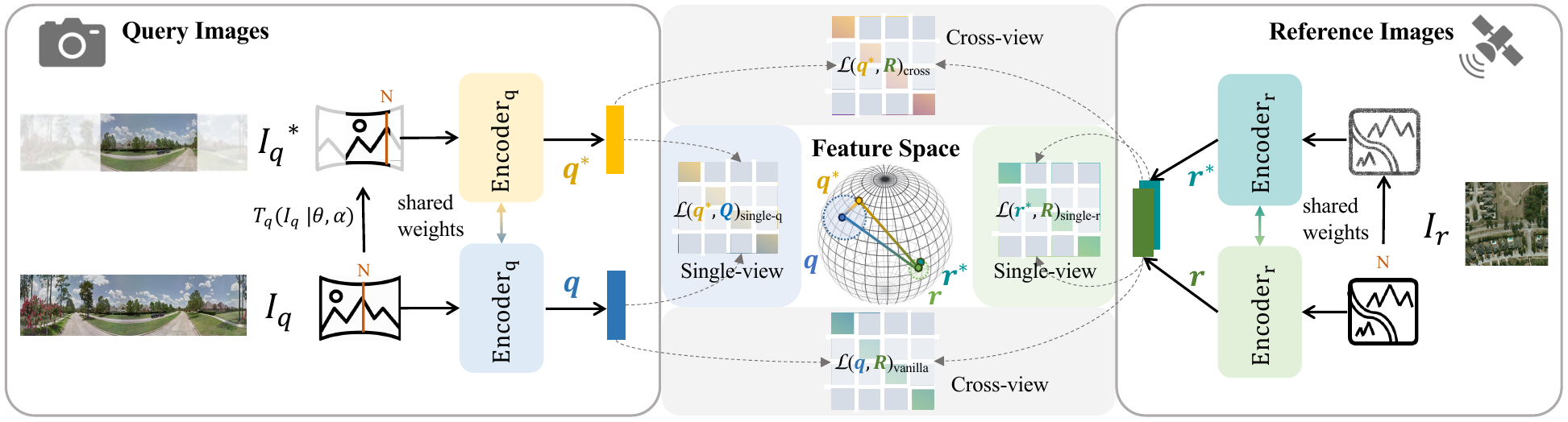}
    \caption{\textbf{ConGeo's learning pipeline.} For feature representation in the left and right boxes, the North-aligned ground image ($I_q$), the transformed ground image ($I^{*}_q$), and the aerial view ($I_r$) are sent to their respective encoders. Then in the feature space, the single- and cross-view contrastive learning losses are applied to enforce the proximity of the paired images.}
    \label{fig:method}
\end{figure*}

To obtain robust representation across diverse ground view variations, we introduce ConGeo, a contrastive learning objective that aims to enhance the robustness of geo-localization models 
by enforcing proximity between ground view variations and their original representations. The proposed ConGeo is model-agnostic and can be a learning objective for different base models. Here we use Sample4Geo~\cite{sample4geo_2023_iccv} as the base model to illustrate the proposed method (Fig.~\ref{fig:method}).

\subsection{Overview}
\noindent \textbf{Problem Statement.} Let $\{I_q\}$ be a set of query images (North-aligned ground view), and let $\{I_r\}$ be a set of reference images (aerial view). For geo-localization under arbitrary orientation and limited FoVs, the query image $I_q$ can be cyclically shifted with a random angle and restricted by a specific FoV (\textit{e.g.}, 70$^{\circ}$, 90$^{\circ}$, 180$^{\circ}$, and 360$^{\circ}$), resulting in its variation $I^{*}_q$ that still shows the same location. For each $I_q$ or $I^{*}_q$, the cross-view geo-localization aims to retrieve the corresponding reference image in $\{I_{r}\}$. 

\noindent \textbf{Architecture.} The classic architecture for geo-localization consists of a siamese network, based on Convolutional Neural Network (CNN) or Vision Transformer (ViT) for feature encoding~\cite{sample4geo_2023_iccv,transgeo_cvpr_2022,safa_nips_2019,jointloc_2020_cvpr}. More specifically, a set of query features $Q:=\{q\}$ is obtained by passing query images into the query encoder: $q=\mathrm{Encoder}_{\mathrm{q}}(I_q)$. Similarly, the set of reference features $R: =\{r\}$ is obtained by $r=\mathrm{Encoder}_{\mathrm{r}}(I_r)$. To boost the model's robustness, ConGeo includes ground image variations $\{I_q{^*}\}$. Such variations are obtained by applying a transformation $T$ to each ground view $I_q$. A random orientation angle $\theta$ is first used to horizontally shift the ground image, followed by the application of a FoV angle $\alpha$ to crop the panorama query to a FoV query $I_q{^*}=T_{q}(I_q | \theta, \alpha)$.

\noindent \textbf{Model Training and Inference.} During training, the dual encoders are learned according to the ConGeo learning objectives (see below). During inference, depending on the setting (North-aligned, unknown orientation, and limited FoV), the specific query image is processed by the ground view encoder. Then, a set of aerial reference images is ranked as retrieval results based on the cosine similarity between the features of the query and reference images.

\subsection{Learning Objectives}

\textbf{Single-view Contrastive Learning.}
To enhance the consistency in feature representations between views across orientations and FoV variations, a single-view contrastive learning objective is designed to produce similar representations between the original images (North-aligned, full panorama) and their transformed counterpart.
The loss for the ground view is computed as:
\begin{equation}
\label{intra}
    \mathcal{L}(q^{*}, Q)_{\text{single-q}} = -\log \frac{\exp\left(q^{*} \cdot q_{+} / \tau_q\right)}{\sum_{q_i \in Q} \exp\left(q^{*} \cdot q_i / \tau_q\right)},
\end{equation}
where $q_i$, $q^{*}$ are the feature embeddings of the original and transformed query image respectively, $q_{+}$ denotes the positive one corresponding to $q^{*}$.

A similar single-view contrastive loss is also applied to the aerial views. However, unlike the ground view contrastive loss that is applied between ground view variations, the aerial view contrastive loss is designed to enforce the similarity between two distorted versions of the same reference image using random data augmentations, aiming to obtain more robust feature representations~\cite{chen2020simclr}. The aerial view contrastive objective is:
\begin{equation}
\label{intra_r}
    \mathcal{L}(r^{*}, R)_{\text{single-r}} = -\log \frac{\exp\left(r^{*} \cdot r_{+} / \tau_r\right)}{\sum_{r_i \in R} \exp\left(r^{*} \cdot r_i / \tau_r\right)},
\end{equation}
where $r_i$ is the feature embedding of the original aerial image, $r_{+}$ denotes the positive one corresponding to $r^{*}$, which is the feature of another possible augmentation of the same aerial image. $\tau_r$ and $\tau_q$ are learnable temperature parameters~\cite{sample4geo_2023_iccv}.

\noindent \textbf{Cross-view Contrastive Learning.}
The cross-view contrastive learning in ConGeo comprises two alignment objectives: First, the loss of the base method is used\footnote{Thanks to its flexible learning objective, ConGeo can be plugged into other existing geo-localization models. In this section, we use Sample4Geo~\cite{sample4geo_2023_iccv} as an example. In the experiments, we also plug ConGeo into TransGeo~\cite{transgeo_cvpr_2022}, and SAIG-D~\cite{zhu2023simplesaigd}.}, which we refer to as the vanilla loss; it is a cross-view alignment loss, where the model learns to match the embeddings of a query image $q$ with the embeddings of its corresponding aerial image $r \in R$. In Sample4Geo~\cite{sample4geo_2023_iccv}, for example, the vanilla loss is:
\begin{equation}
\label{infonce}
    \mathcal{L}(q, R)_{\text {vanilla}}=-\log \frac{\exp \left(q \cdot r_{+} / \tau_v\right)}{\sum_{r_i\in R}\exp \left(q \cdot r_i / \tau_v\right)},
\end{equation}
where $R$ is a set of reference images, and $r_+$ is the reference feature embedding that matches the query embedding.

Second, to further reinforce the model's robustness to ground view variations, we enforce the alignment between transformed ground images and aerial images with a cross-view contrastive objective:
\begin{equation}
\label{inter}
    \mathcal{L}(q^{*}, R)_{\text{cross}} = -\log \frac{\exp\left(q^{*} \cdot r_{+} / \tau_c\right)}{\sum_{r_i\in R} \exp  \left(q^{*} \cdot r_i / \tau_c\right)},
\end{equation}
with $\tau_v$ and $\tau_c$ as learnable temperature parameters.

\noindent \textbf{Final Loss.}
Combining all the terms previously described, the total loss of ConGeo can be written as:
\begin{equation}
\label{final}
    \mathcal{L} = \mathcal{L}_{\text {vanilla}} + w_{1} \mathcal{L}_{\text {single-q}} + w_{2} \mathcal{L}_{\text{single-r}} + w_{3}\mathcal{L}_{\text {cross}},
\end{equation}
where $w_1$, $w_2$ and $w_3$ are factors balancing the contribution of the different learning objectives.

\section{Experiments}

\subsection{Datasets and Evaluation Metrics}
\textbf{Datasets.} Experiments are performed on four CVGL datasets.  
\textbf{CVUSA}~\cite{cvusa_cvpr_2017} contains 35,532 view pairs for training and 8,884 for evaluation. \textbf{CVACT}~\cite{lending_2019_CVPR} is split into training, validation, and test sets, where the first two are of the same size as CVUSA while the test set has 92,802 image pairs. \textbf{VIGOR}~\cite{vigor_cvpr_2021} contains 90,618 satellite images and 105,214 street-view images and can be divided into two train-test splits: same-area and cross-area. The training and test data in the cross-area subset were collected from different cities. \textbf{University-1652}~\cite{zheng2020university} contains drone, satellite and street view images. The street view images are commonly with unknown orientation and limited FoV, enabling a real-world evaluation under this challenging setting. Detailed dataset descriptions can be found in the supplementary material.

\noindent \textbf{Evaluation Metrics.} Following the literature~\cite{jointloc_2020_cvpr, sample4geo_2023_iccv, zhu2023simplesaigd}, we use Top-$k$ recall, denoted as R@$k$, to measure the retrieval performance. For each ground view query image, the aerial images are ranked based on cosine similarity with ground view representation. If the ground-truth aerial image is among the top~$k$ retrieved images, the retrieval is considered a success under R@$k$. As in previous methods~\cite{zheng2020university, sample4geo_2023_iccv}, we also use Average Precision (AP) for one-to-many and many-to-one matching in the University-1652 dataset.

\subsection{Implementation Details}

\noindent \textbf{Base Models and Data Preprocessing.} 
Unless specified, ConGeo uses Sample4Geo~\cite{sample4geo_2023_iccv} as the base model, with ConvNeXt-B~\cite{convnext_cvpr_2022} as the backbone. In Section~\ref{sec:plugin}, we also plug ConGeo into TransGeo~\cite{transgeo_cvpr_2022} and SAIG-D~\cite{zhu2023simplesaigd}. We follow the default settings and data augmentation methods used in those models.

\noindent \textbf{Hyper-parameters and Environment.} Loss weights $w_1$, $w_2$, and $w_3$ are empirically set to 0.5, 0.5, and 0.25, corresponding to a higher weight on the single-view contrastive loss and lower weight on the cross-view one. For training, the orientation angle $\theta$ is randomly drawn between 0$^{\circ}$ and 360$^{\circ}$ while the FoV angle $\alpha$ is set to 180$^{\circ}$. For evaluation, the orientation angle and FoV angle depend on different settings (see Section~\ref{settings}). Experiments comparing different hyper-parameters are provided in the supplementary materials. Besides, following the default setting, for those who used Sample4Geo as the base model, the weights of the query encoder and the reference encoder are shared. For each experiment, the model is trained for 60 epochs with a batch size of 16. We use the AdamW optimizer with an initial learning rate of 0.0001 and a cosine learning rate scheduler. A single NVIDIA GeForce RTX 4090 is used for all the experiments.

\subsection{Experimental settings}
\label{settings}
The evaluation is performed for three different settings.

\noindent \textbf{ - North-aligned.} This is the common configuration of the task: the full-view North-aligned ground image is used to retrieve the aerial view. 

\noindent \textbf{ - Unknown Orientation.} When testing, we shift the full-view ground image by a random angle and use it as the query. This setting is also noted as FoV=360$^{\circ}$.

\noindent \textbf{ - Limited FoV.} We first apply a random cyclical shift to the ground image (same as unknown orientation) and then reduce the FoV to 70$^{\circ}$, 90$^{\circ}$, or 180$^{\circ}$.

For the CVUSA and CVACT datasets, we evaluate the model's performance under these 3 settings. For the VIGOR dataset, we use the unknown orientation and limited FoV settings to verify the method's cross-location robustness. Furthermore, we use the University-1652 dataset for Street-to-Satellite (St2S, many-to-one) and Satellite-to-Street (S2St, one-to-many) retrieval, showing the robustness for street view images. We also test the models' generalization ability on four unseen ground view variations on the CVUSA dataset.

\begin{table*}[t]
\small
\setlength{\tabcolsep}{1.5pt}
\centering
\resizebox{\linewidth}{!}{
\begin{tabular}{c | l | c  c c c c c | c c c c c | c c c c c | c c c c | c}
\toprule
 \multirow{2}{*}{Set} & \multirow{2}{*}{Methods} & \multicolumn{6}{c|}{FoV=$360^{\circ}$}   & \multicolumn{5}{c|}{FoV=$180^{\circ}$}     & \multicolumn{5}{c|}{FoV=$90^{\circ}$}   & \multicolumn{4}{c|} {FoV=$70^{\circ}$} & Avg. \\ 
&  &  &  R@1   &  R@5   &  R@10  &  R@1\% & &  R@1   &  R@5   &  R@10  &  R@1\% &  &  R@1   &  R@5   &  R@10  &  R@1\% & &  R@1   &  R@5   &  R@10  &  R@1\% & R@1 \\
\hline  
& \footnotesize CVFT \cite{shi2020optimalCVFT}  &   & 23.4 & 44.4 & 55.2 & 86.6 & & 8.1 & 24.3 & 34.5 & 75.2  & &  4.8   & 14.8 & 23.2 & 61.2 &  & 3.8 & 12.4 & 19.3 & 55.6 & 10.0 \\ 
& \footnotesize SEH \cite{guo2022softSEH} & & 85.4 & 93.5 & 95.8 & - & & 53.7 & 72.3 & 79.0 & - & & 16.6 & 32.2 & 40.3 & - & & 7.8 & 18.8 & 25.6 & - & 40.9 \\
& \footnotesize DSM \cite{jointloc_2020_cvpr}  & & 78.1 & 89.5 & 92.9 & 98.5  &  & 48.5 & 68.5 & 75.6 & 93.0 & & 16.2 & 31.4 & 39.9 & 71.1 & & 8.8 & 19.9 & 27.3 & 61.2 & 37.9\\
& \footnotesize  SAIG-D \cite{zhu2023simplesaigd} & & 72.0 & 90.2 & 94.0 & 99.1 &  & 52.5 & 78.1 & 85.8 & 97.7 & & 26.7 & 50.2 & 59.8 & 86.6 & & 20.9 & 41.4 & 51.2 & 80.4 & 43.0 \\
& \footnotesize Sample4Geo \cite{sample4geo_2023_iccv} & & 93.3 & 97.5 & 98.0 & 99.1 & & 84.6 & 95.9 & 97.6 & 99.5 & & 55.1 & \textbf{78.3} & \textbf{85.0} & \textbf{96.6} & & 40.9 & 65.4 & 74.1 & 93.0 & 68.5 \\
& \footnotesize \textbf{ConGeo} & & \textbf{96.6} & \textbf{98.9} & \textbf{99.2} & \textbf{99.7} & & \textbf{92.3} & \textbf{97.9} & \textbf{98.7} & \textbf{99.7} & & \textbf{55.5} & 75.4 & 81.5 & 93.9 & & \textbf{49.1} & \textbf{70.8} & \textbf{78.0} & \textbf{93.1} & \textbf{73.4}\\
\cline{2-22}
\rowcolor{mygrey} \cellcolor{white} & \footnotesize Sample4Geo\cite{sample4geo_2023_iccv} & & 16.3 & 26.1 & 31.4 & 51.7 & & 4.1 & 8.4 & 11.3 & 30.4 & & 2.5 & 6.7 & 9.8 & 26.7 & & 1.5 & 4.6 & 6.7 & 20.4 & 6.1\\
\rowcolor{mygrey} \cellcolor{white} & \footnotesize Sample4Geo† \cite{sample4geo_2023_iccv}  & & \textbf{93.2} & \textbf{98.2} & \textbf{99.0} & \textbf{99.8} & & 84.6 & 95.9 & 97.6 & 99.5 & & 45.1 & 64.8 & 71.3 & 86.5 & & 28.4 & 47.1 & 54.9 & 75.8 & 62.8 \\
\rowcolor{mygrey} \cellcolor{white} \multirow{-9}{*}{\rotatebox{90}{CVUSA}} & \footnotesize \textbf{ConGeo}  & & 85.2 & 95.1 & 96.9 & 98.9 &  & \textbf{92.3} & \textbf{97.9} & \textbf{98.7} & \textbf{99.7} & & \textbf{55.9} & \textbf{73.2} & \textbf{79.0} & \textbf{90.9} & & \textbf{37.1} & \textbf{55.7} & \textbf{62.8} & \textbf{81.4} & \textbf{67.6}\\

\hline
& \footnotesize CVFT \cite{shi2020optimalCVFT}   &  & 26.8 & 46.9 & 55.1 & 81.0 & & 7.1 & 18.5 & 26.8 & 63.9   &  & 1.9 & 6.3 & 10.5 & 39.3  &  & 1.5 & 5.1 & 8.2 & 34.6  & 9.3 \\ 
& \footnotesize SEH \cite{guo2022softSEH} & & 77.4 & 88.6 & 90.9 & - & & 47.7 & 67.9 & 74.3 & - & & 13.9 & 28.4 & 36.2 & - & & 6.9 & 16.5 & 22.3 & - & 36.5\\
& \footnotesize DSM \cite{jointloc_2020_cvpr}  & & 72.9  & 85.7  & 88.9   & 95.3 & & 49.1  & 67.8     & 74.2     & 89.9 & & 18.1  & 33.3  & 40.9  & 68.7  & & 8.3  & 20.7     & 27.1     & 57.1  & 37.1 \\
& \footnotesize Sample4Geo \cite{sample4geo_2023_iccv} &  & 82.4 & \textbf{90.6} & 92.3 & 95.4 & & 58.9 & 79.8 & 85.3 & 95.0 & & 27.9 & 52.0 & 62.3 & \textbf{87.0} & & 18.8 & 40.4 & 51.0 & \textbf{81.3} & 47.0 \\
&  \footnotesize \textbf{ConGeo} &  & \textbf{83.0} & \textbf{90.6} & \textbf{92.4} & \textbf{96.3} & & \textbf{70.3} & \textbf{85.2} & \textbf{88.6} & \textbf{95.1} & & \textbf{40.6} & \textbf{62.6} & \textbf{69.8} & 86.6 & & \textbf{24.6} & \textbf{45.3} & \textbf{54.3} & 80.6 & \textbf{54.6} \\
\cline{2-22}
\rowcolor{mygrey} \cellcolor{white} & \footnotesize Sample4Geo\cite{sample4geo_2023_iccv} & & 12.6 & 17.5 & 19.9 & 31.2 & & 3.4 & 8.0 & 10.6 & 24.5 & & 1.9 & 5.5 & 7.9 & 23.9 & & 1.0 & 3.2 & 5.0 & 16.8 & 4.7 \\
\rowcolor{mygrey} \cellcolor{white}  & \footnotesize Sample4Geo†\cite{sample4geo_2023_iccv} & & \textbf{77.1} & \textbf{89.8} & \textbf{92.6} & \textbf{97.2} & & 58.9 & 79.8 & 85.3 & 95.0 & & 22.4 & 44.0 & 52.7 & 78.3 & & 12.5 & 28.5 & 37.5 & 67.4 & 42.7 \\
\rowcolor{mygrey} \cellcolor{white} \multirow{-8}{*}{\rotatebox{90}{CVACT}}  &  \footnotesize \textbf{ConGeo} & & 62.6 & 79.9 & 84.7 & 93.9 & & \textbf{70.3} & \textbf{85.2} & \textbf{88.6} & \textbf{95.1} & & \textbf{34.8} & \textbf{56.9} & \textbf{64.8} & \textbf{83.6} & & \textbf{18.5} & \textbf{37.4} & \textbf{46.7} & \textbf{72.5} & \textbf{46.6} \\
\bottomrule
\end{tabular}}
\caption{\small Comparison on the unknown orientation setting and limited FoV setting on CVUSA~\cite{cvusa_cvpr_2017} and CVACT Val~\cite{jointloc_2020_cvpr} datasets. ``-'' means the score is not provided in the original paper. The best performance is in \textbf{bold}. Results of models trained with FoV-specific images are reported in the white background, while results from a single model without FoV-specialized training are reported with a \colorbox{mygrey}{grey background}. † denotes using the same training FoV as ConGeo.}
\label{tab: fov}
\end{table*}

\section{Results}

\subsection{Performance under Different Settings}
\label{sec:different_setting}
As in previous works~\cite{jointloc_2020_cvpr}, we assess the effectiveness of ConGeo in the unknown orientation setting and the limited FoV settings on the CVUSA and CVACT validation datasets in Table~\ref{tab: fov}.
Unlike previous methods~\cite{jointloc_2020_cvpr, yang2021crossl2ltr, zhu2023simplesaigd}, which train different models for each setting to improve performance, the single model trained with ConGeo excels in all settings. 
As shown in the rows highlighted in grey in Table~\ref{tab: fov}, ConGeo improves the base model's (Sample4Geo) performance by a large margin (68.9\% and 53.4\% R@1 improvement for FoV=360$^{\circ}$ and FoV=90$^{\circ}$, respectively) and significantly outperforms the state-of-the-art methods on most of the challenging settings on the CVUSA dataset, \textit{without training models on separate FoVs}. 
Besides, when training ConGeo on each FoV separately, the model's performance can be further enhanced in different settings, significantly surpassing all competitors on all FoVs. 

Similarly, ConGeo outperforms the state-of-the-art model's R@1 under 360$^{\circ}$, 180$^{\circ}$, 90$^{\circ}$, and 70$^{\circ}$ settings on the CVACT validation set by 0.6\%, 11.4\%, 12.7\%, and 5.8\% respectively. ConGeo also shines on this dataset with a single model, surpassing the previous approaches in most settings by a substantial margin.
In short, these results show the consistent superiority of ConGeo over previous state-of-the-art methods and its robustness in handling ground view variations.

\begin{figure}[t]
    \begin{minipage}{0.5\linewidth}
        \centering
        \resizebox{0.98\textwidth}{!}{ 
            \begin{tabular}{l|cc|cc|cc}
            \toprule
            \multirow{2}{*}{Methods} & \multicolumn{2}{c|}{CVUSA} & \multicolumn{2}{c|}{CVACT Val}& \multicolumn{2}{c}{CVACT Test}  \\
            & R@1 & R@1\% &  R@1  & R@1\%&  R@1 & R@1\%\\ \hline
            CVFT \cite{shi2020optimalCVFT} & 61.4 & 99.0 & 61.1 & 95.9 & - & - \\
            DSM \cite{jointloc_2020_cvpr}  & 92.0 & 99.7 & 82.5 & 97.3 & - & - \\
            TransGeo~\cite{transgeo_cvpr_2022} & 94.1 & \underline{99.8} & 85.0 & 98.4 & - & -\\
            GeoDTR~\cite{zhang2023crossGeoDTR} & 95.4 & \textbf{99.9} & 86.2 &  \underline{98.8} & 64.5 & \textbf{98.7}\\ 
            SAIG-D~\cite{zhu2023simplesaigd} & 96.3 & \textbf{99.9} & 89.1 &  \textbf{98.9} & 67.5 & 96.8\\ 
            \rowcolor{mygrey} Sample4Geo~\cite{sample4geo_2023_iccv} & \textbf{98.7} & \textbf{99.9} & \textbf{90.8} & \underline{98.8}  & \underline{71.5}  & \textbf{98.7}\\
            \rowcolor{mygrey} \textbf{ConGeo} & \underline{98.3}  & \textbf{99.9} & \underline{90.1} & 98.2 & \textbf{71.7} & \underline{98.3}\\
            \bottomrule
            \end{tabular}
        }
        \captionof{table}{\small Comparison of the North-aligned setting on CVUSA and CVACT datasets. The second-best performance is \underline{underlined}. ``-'' means the score is not provided in the original paper.}
        \label{tab:standard}
    \end{minipage}
    \hfill
    \begin{minipage}{0.45\linewidth}
        \centering
        \includegraphics[width=0.9\linewidth]{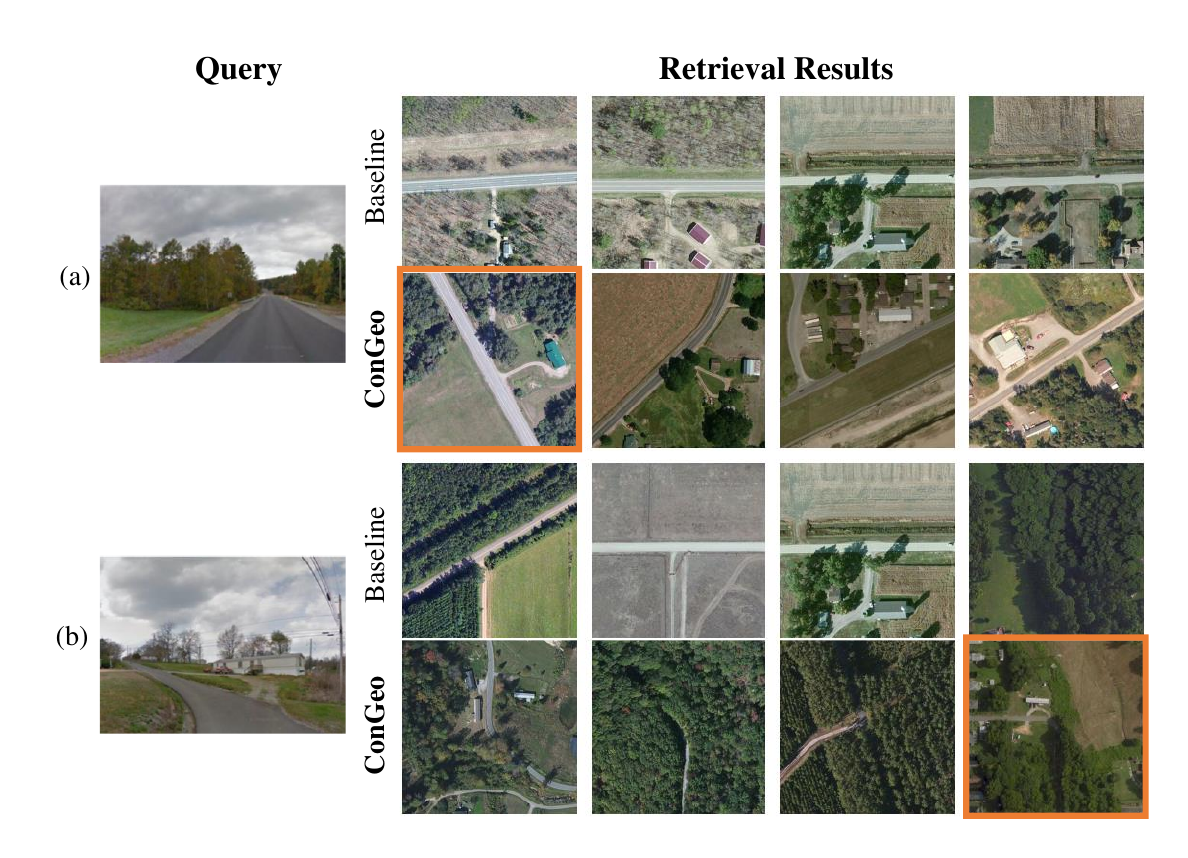}
        \caption{Examples of the top-4 retrieved images from ConGeo and the baseline when FoV=90$^{\circ}$. Images in the orange box denote the correct results.} 
        \label{fig:examples}
    \end{minipage}
\end{figure}

Moreover, ConGeo maintains competitive performance in the North-aligned setting. As shown in Table~\ref{tab:standard}, ConGeo always ranks within the top two for R@1 both on the CVUSA and CVACT datasets. Moreover, by comparing Table~\ref{tab:standard} and Table~\ref{tab: fov}, we observe that when confronting challenging settings, baselines experience a considerable performance drop (as seen from the results with grey background in both tables). For example, for a query image with FoV=180$^{\circ}$, R@1 of the base model drops by 82.4\%; while shifting the North-aligned ground view with unknown orientation leads to an R@1 decrease of 94.6\%. On the contrary, ConGeo maintains robustness to various ground view shifts with R@1 dropping by only 13.1\% for 360$^{\circ}$ and by 6.0\% for 180$^{\circ}$.

Fig.~\ref{fig:examples} shows two examples of top-4 retrieved images when the query image's FoV=90$^{\circ}$. Limited FoV images offer restricted semantic and spatial information and thus are more challenging. In Fig.~\ref{fig:examples}~(a), the baseline model only retrieves aerial images with a horizontal road, indicating that the model heavily relies on spatial correspondence; ConGeo's results are more diverse among road directions. Fig.~\ref{fig:examples}~(b) shows a building next to a half-circle side road. Several images retrieved by ConGeo include similar visual features, showing that ConGeo elicits the model's focus on information that is consistent across ground view variations.

\subsection{Ablations}

We perform two ablation studies to show the advantages of ConGeo over alternative data augmentations and the impact of each component of ConGeo. 

\noindent \textbf{Comparison with Data Augmentations} (Table~\ref{tab:aug}). One possible way to improve robustness under view variations is through data augmentation. We compare ConGeo with three methods: ``Shift'' (random cyclical shift), ``FoV'' (random query image FoV cropping, from 70$^{\circ}$ to 360$^{\circ}$), and ``Rotate'' (random aerial images rotation by 90$^{\circ}$, 180$^{\circ}$ or 270$^{\circ}$). 
Results show that these methods can improve the base model under unknown orientation and limited FoV. However, their overall performance stays far lower than ConGeo, especially since their performance in the North-aligned setting notably decreases, showing the superiority of contrastive objectives over targeted data augmentation.
Compared to ConGeo, all the data augmentation approaches lack a crucial single-view contrastive objective, which explicitly enforces \textit{discovering the joint features shared by different ground view variations} and therefore being robust to them. 
\begin{table}[t]
    \centering
        \resizebox{0.99\columnwidth}{!}{ 
            \begin{tabular}{c c c| c c c c | c c c c |c c c c | c c c c| c }
            \toprule
            \multicolumn{3}{c|}{Aug. type} & \multicolumn{4}{c|}{North-aligned} & \multicolumn{4}{c|}{FoV=360$^{\circ}$} & \multicolumn{4}{c|}{FoV=180$^{\circ}$} & \multicolumn{4}{c|}{FoV=90$^{\circ}$} & Avg.  \\
            Shift & FoV & Rotate & R@1 &  R@5 & R@10 & R@1\% & R@1 & R@5 & R@10 & R@1\% & R@1 & R@5 & R@10 & R@1\% & R@1 & R@5 & R@10 & R@1\% & R@1 \\
            \hline
              & & & \textbf{98.7} & \textbf{99.7} & \textbf{99.8} & \textbf{99.9} & 16.3 & 26.1 & 31.4 & 51.7 & 4.1 & 8.4 & 11.3 & 30.4 & 2.5 & 6.7 & 9.8 & 26.7 & 30.4 \\
             \checkmark & & & 94.4 & 98.1 & 99.6 & 93.2 & \textbf{93.1} & \textbf{97.6} & \textbf{98.2} & \underline{99.1} & \underline{73.8} & \underline{88.0} & 90.9 & 95.8 & 35.1 & 54.2 & 61.2 & 77.6 & \underline{74.1}\\ 
             \checkmark & \checkmark &  & 90.3 & 97.3 & 98.4 & \underline{99.6} & 84.1 & 95.0 & 96.9 & \textbf{99.4} & 63.6 & 84.6 & 90.1 & \underline{98.2} & 32.2 & 55.1 & 64.5 & \underline{87.4} & 67.6\\ 
             \checkmark & \checkmark & \checkmark & 88.9 & 96.0 & 97.1 & 99.0 & \underline{89.0} & \underline{96.1} & \underline{97.3} & 98.9 & 71.8 & 87.9 & \underline{91.3} & 97.2 & \underline{39.3} & \underline{59.9} & \underline{67.9} & 85.7 & 72.3\\ 
             \hline
            \multicolumn{3}{c|}{ConGeo} & \underline{98.3} & \underline{99.6} & \underline{99.7} & \textbf{99.9} & 85.2 & 95.1 & 96.9 & 98.9 & \textbf{92.3} & \textbf{97.9} & \textbf{98.7} & \textbf{99.7} & \textbf{55.9} & \textbf{73.2} & \textbf{79.0} & \textbf{90.9}  & \textbf{82.9} \\
            \bottomrule
            \end{tabular}
        }   
        \caption{\small The comparison of results between ConGeo and task-specific data augmentations. We incorporate augmentations into Sample4Geo, ``Shift'' denotes using shifted query images and ``FoV'' denotes using query images of limited FoVs, ``Rotate'' randomly rotating aerial images with an angle in \{90$^{\circ}$, 180$^{\circ}$, 270$^{\circ}$\} as data augmentation. The second-best performance is \underline{underlined}. }
        \label{tab:aug}
\end{table}

\noindent \textbf{Ablations of Loss Components} (Table~\ref{tab:ablation1}).
The aerial view contrastive objective ($\mathcal{L}_{\text{single-r}}$) slightly enhances the performance (Row 1 and 2), which serves as the basis for the following ablations. The ground view contrastive loss ($\mathcal{L}_{\text{single-q}}$) gradually boosts performance under FoVs as contrastive targets between ground views are added (Row 1, 3, and 4). Finally, adding the cross-view contrastive loss on top of single-view losses yields a notable improvement (Row 4 and 7). 
This indicates that the contrastive learning between the original and shifted ground images plays an essential role in assisting cross-view alignment.

\begin{table}[t]
    \begin{minipage}{0.49\linewidth}
        \centering
        \renewcommand{\arraystretch}{1.1}
        \resizebox{1.00\columnwidth}{!}{ 
    \begin{tabular}{c| c | c c| c c | c  c | c c}
    \toprule
    \multirow{2}{*}{No.} & \multirow{2}{*}{$\mathcal{L}_{\text{single-r}}$} &\multicolumn{2}{c|}{$\mathcal{L}_{\text{single-q}}$} & \multicolumn{2}{c|}{$\mathcal{L}_{\text{cross}}$}  & \multicolumn{2}{c|}{FoV=180$^{\circ}$}  & \multicolumn{2}{c}{FoV=90$^{\circ}$} \\
     &  & Shift & FoV & Shift & FoV &  R@1 & R@1\% & R@1 & R@1\%\\
    \hline
     1 & & &   & & & 4.1 & 30.4 & 2.5 & 26.7 \\ 
     2 & \checkmark & &   & & & 15.7 & 60.0 & 7.7  & 43.3  \\ 
     3 & \checkmark & \checkmark & & & & 44.1 & 93.4 & 17.8 & 73.0\\
     4 & \checkmark & \checkmark & \checkmark & &  & 37.9 & 81.2 & 20.5 & 59.9 \\
     5 & \checkmark &  & & \checkmark & \checkmark & 91.5 & \textbf{99.7} & 40.2 &  89.0\\ 
     6 & \checkmark & \checkmark &  & \checkmark &  & 81.7 & 98.8 & 35.8 &  81.6 \\
     7 & \checkmark & \checkmark & \checkmark & \checkmark & \checkmark & \textbf{92.3} & \textbf{99.7} & \textbf{55.9} & \textbf{90.9} \\
    \bottomrule
    \end{tabular}}   
    \caption{Ablation studies on FoV=180$^{\circ}$ and FoV=90$^{\circ}$ on the CVUSA dataset. ``Shift'' and ``FoV'' mean cyclic shift and FoV cropping for ground view images. }
    \label{tab:ablation1}
    \end{minipage}
    \hfill
    \begin{minipage}{0.5\linewidth}
    \centering
        \resizebox{0.95\columnwidth}{!}{ 
            \begin{tabular}{c| c c | c c| c c }
            \toprule
            \multirow{2}{*}{Methods}  & \multicolumn{2}{c|}{FoV=360$^{\circ}$} & \multicolumn{2}{c|}{FoV=180$^{\circ}$} & \multicolumn{2}{c}{FoV=90$^{\circ}$} \\
            &  R@1 & R@1\% & R@1 & R@1\% & R@1 & R@1\% \\
            \hline
            TransGeo & 13.5 & 59.4 & 4.5 & 42.2 & 0.4 & 13.9\\ 
            TransGeo + DA & \textbf{75.9} & \textbf{99.2} & 47.8 & 94.9 & 18.7 & 74.7 \\ 
            \textbf{ConGeo}[TransGeo] & 52.7 & 97.2 & \textbf{54.8} & \textbf{97.4} & \textbf{26.9} & \textbf{83.8}\\ 
            \hline
            SAIG-D & 12.5  & 69.4 & 3.3 & 40.1 & 0.3 & 10.2 \\ 
            SAIG-D + DA & 64.8 & 98.6 & 49.3 & 96.9 & \textbf{29.7} & \textbf{90.0} \\ 
            \textbf{ConGeo}[SAIG-D] & \textbf{70.7} & \textbf{98.9} & \textbf{54.9} & \textbf{97.3} & 24.0 & 80.4\\ 
            \hline
            Sample4Geo &  16.3 & 51.7 & 4.1 & 30.4 & 2.5 & 26.7 \\
            Sample4Geo + DA & 84.1 & \textbf{99.4} & 63.6 & 98.2 & 32.2 & 87.4 \\ 
            \textbf{ConGeo}[Sample4Geo] & \textbf{85.2}  & 98.9 & \textbf{92.3} & \textbf{99.7} & \textbf{55.9} & \textbf{90.9}\\
            \bottomrule
            \end{tabular}
        }   
        \caption{\small ConGeo with three base models on the CVUSA dataset including TransGeo~\cite{transgeo_cvpr_2022}, SAIG-D~\cite{zhu2023simplesaigd} and Sample4Geo~\cite{sample4geo_2023_iccv}. Note that ``DA'' means data augmentation.}
        \label{tab:plugin}
    \end{minipage}
\end{table}

\subsection{Adaptability to Different Base Models}
\label{sec:plugin}
ConGeo is model-agnostic: it can be plugged into different CVGL systems and boosts their robustness to ground view variations. To demonstrate this, we choose two representative methods, TransGeo \cite{transgeo_cvpr_2022} and SAIG-D \cite{zhu2023simplesaigd} as base models, besides Sample4Geo. For each base model, we follow the default configurations --- including backbone and data augmentations --- and use the vanilla loss in the base model instead of Eq.\eqref{final}. We train the ConGeo-augmented models on all FoVs jointly (same as the results with grey background in Table~\ref{tab: fov}). We compare each model with two baselines: (1) the base model trained on North-aligned data, (2) the base model trained with strong data augmentations, including shifting each ground image with a random angle and cropping it with a FoV between 70$^{\circ}$ and 360$^{\circ}$ (referred to as ``DA''). 
As shown in Table \ref{tab:plugin}, adding ConGeo improves ViT-based TranGeo's R@1\% performance by 37.8\% and 69.9\% for FoV=360$^{\circ}$ and FoV=90$^{\circ}$, respectively. SAIG-D combines convolutional stem and self-attention layers in the encoder and the improvement with ConGeo is also remarkable. ConGeo-augmented models also outperform strong data augmentations under most FoVs.
In summary, ConGeo can be plugged into three different baselines with both CNN-based and ViT-based backbones, demonstrating its versatility and effectiveness across model architectures. 

\subsection{Robustness across Other Ground View Variations}
\label{sec:other_ground_view}
\noindent \textbf{Robustness across Locations} (Table~\ref{tab:cross}).
In cross-view geo-localization, the VIGOR dataset is particularly challenging, because the two views are not center-aligned. In particular, VIGOR's cross-area subset is regarded as a standard benchmark to test the model's robustness to data across different locations, since training and test images are taken from different areas~\cite{vigor_cvpr_2021, xia2022visual, sample4geo_2023_iccv}. As shown in Table~\ref{tab:cross}, ConGeo outperforms the baselines on both the cross-area and same-area subsets. 
Note that we report the performance of a single model for all FoVs, for ConGeo and Sample4Geo (grey background). The results indicate that ConGeo consistently improves the model's robustness on cross-area data.

\begin{table}[t]
    \begin{minipage}{0.55\linewidth}
        \centering
        \resizebox{0.95\columnwidth}{!}{ 
        \begin{tabular}{c| c c | c c | c c| c c}
        \toprule
        \multirow{3}{*}{Methods} & \multicolumn{4}{c|}{Cross-Area} & \multicolumn{4}{c}{Same-Area} \\
        & \multicolumn{2}{c|}{FoV=360$^{\circ}$}  & \multicolumn{2}{c|}{FoV=90$^{\circ}$} & \multicolumn{2}{c|}{FoV=360$^{\circ}$}  & \multicolumn{2}{c}{FoV=90$^{\circ}$} \\
        &  R@1 & R@1\% & R@1 & R@1\% & R@1 & R@1\% & R@1 & R@1\% \\
        \hline
        VIGOR~\cite{vigor_cvpr_2021} & 1.4 & 44.6 & - & - & 19.1 & 95.1 & - & - \\ 
        TransGeo~\cite{transgeo_cvpr_2022} & 5.5 & 66.9 & - & - & 47.7 & \textbf{99.3} & - & -\\
        \hline
        \rowcolor{mygrey} Sample4Geo~\cite{sample4geo_2023_iccv} & 9.0 & 43.7 & 0.5 & 21.6 & 14.2 & 54.9 & 1.1 & 30.6 \\
         \rowcolor{mygrey}\textbf{ConGeo} & \textbf{16.2} & \textbf{72.9} & \textbf{3.9} & \textbf{54.3} & \textbf{61.9} & 98.4 & \textbf{8.5} & \textbf{68.7} \\ 
        \bottomrule
        \end{tabular}}  
        \caption{\small Comparison on the VIGOR dataset~\cite{vigor_cvpr_2021}. ``Cross-Area'' and ``Same-Area'' mean its cross-area subset and same-area subset, respectively.}
        \label{tab:cross}
    \end{minipage}
    \hfill
    \begin{minipage}{0.43\linewidth}
        \centering
            \resizebox{0.85\columnwidth}{!}{ 
            \begin{tabular}{c |cc| cc} 
        \toprule
        \multirow{2}{*}{Methods} & \multicolumn{2}{c|}{St2S} & \multicolumn{2}{c}{S2St}\\
         & R@1 & AP & R@1 & AP \\
        \hline
        University-1652 \cite{zheng2020university} & 0.6 & 1.6 & 0.9 & 1.0 \\
        LPN~\cite{wang2021eachlpn} & 0.7 & 1.8 & 1.4 & 1.3\\
        Sample4Geo$^{\star}$ \cite{sample4geo_2023_iccv} & 4.9 & 8.1 & 6.6 & 6.1 \\ 
        \textbf{ConGeo} & \textbf{5.9} & \textbf{9.2} & \textbf{6.8} & \textbf{6.4}\\
        \bottomrule
    \end{tabular}
    }
   \caption{\small Comparison on University-1652~\cite{zheng2020university}. $^{\star}$ denotes the reproduced model since the pre-trained weights are not provided.}
\label{tab:uni}
    \end{minipage}
\end{table}
\begin{table}[t]
    \centering
        \resizebox{0.96\columnwidth}{!}{ 
            \begin{tabular}{c| c c c c | c c c c |c c c c | c c c c | c}
            \toprule
            \multirow{2}{*}{Methods} & \multicolumn{4}{c|}{Random FoVs} & \multicolumn{4}{c|}{Random Zooming} & \multicolumn{4}{c|}{Gaussian Noise} & \multicolumn{4}{c|}{Motion Blur} & {Avg.}\\
            & R@1 &  R@5 & R@10 & R@1\% & R@1 & R@5 & R@10 & R@1\% & R@1 & R@5 & R@10 & R@1\% & R@1 & R@5 & R@10 & R@1\% & R@1 \\
            \hline
             Sample4Geo & 15.7 & 25.3 & 31.2 & 51.7 & 48.2 & 60.8 & 65.8 & 80.1 & 42.4 & 60.9 & 67.0 & 81.6 & 31.4 & 38.6 & 42.0 & 52.9 & 34.4\\ 
             Sample4Geo + DA & \textbf{85.5} & \textbf{95.1} & 96.6 & 98.8 & 44.5 & 60.6 & 66.7 & 82.7 & 0.2 & 1.0 & 1.7 & 8.8 & 16.8 & 23.2 & 25.9 & 36.0 & 36.8\\ 
            \textbf{ConGeo} & 84.2 & \textbf{95.1} & \textbf{97.0} & \textbf{99.4} & \textbf{68.7} & \textbf{80.3} & \textbf{83.8} & \textbf{92.1} & \textbf{45.8} & \textbf{64.4} & \textbf{70.3} & \textbf{83.4} & \textbf{32.5} & \textbf{40.7} & \textbf{43.5} & \textbf{53.5} & \textbf{57.8}\\
            \bottomrule
            \end{tabular}
        }   
        \caption{\small Comparison on unseen ground view variations between ConGeo and baselines on the CVUSA dataset. Random FoVs are between 0$^{\circ}$ to 360$^{\circ}$, Random Zooming is performed with a ratio between 0.5 and 2.0, Gaussian Noise~\cite{arularasurobust}, and Motion Blur~\cite{arularasurobust} are added with severity: 5.}
        \label{tab:unseen}
\end{table}

\noindent \textbf{Robustness to Street Images} (Table~\ref{tab:uni}). Street images in University-1652 are typical examples of real-world limited FoV images. The St2S and S2St settings pose significant challenges as the limited FoV images can be captured from diverse locations with unknown orientations.
Different from the experiments considering North-aligned panoramas, here we randomly sample two street view images from one location as the inputs to conduct single-view contrastive loss. Detailed descriptions can be found in the supplementary materials. Results in Table~\ref{tab:uni} show that the performance on both settings is improved.

\noindent \textbf{Robustness to Unseen Ground View Variations} (Table~\ref{tab:unseen}). 
In real-world scenarios, the variations of the ground view image are more diverse than orientation and FoV variations. In Table~\ref{tab:unseen}, we evaluate the model across four unseen ground view variations on the CVUSA dataset: Random FoVs, Random Zooming, Gaussian Noise, and Motion Blur. The ConGeo-augmented model exhibits comparable performance to the model with data augmentation on Random FoVs test, while demonstrating significant advantages over the two baselines when tested on other unseen ground view variations. According to Table~\ref{tab:aug} and~\ref{tab:unseen}, the improvements that task-oriented data augmentation brings are \textit{not transferable nor generalizable}, on the contrary, ConGeo represents a way of unleashing the potential of data augmentation by enforcing the learning of invariances.

\begin{figure}[t]
\centering
\begin{minipage}[t]{0.48\textwidth}
\centering
\includegraphics[width=0.98\linewidth]{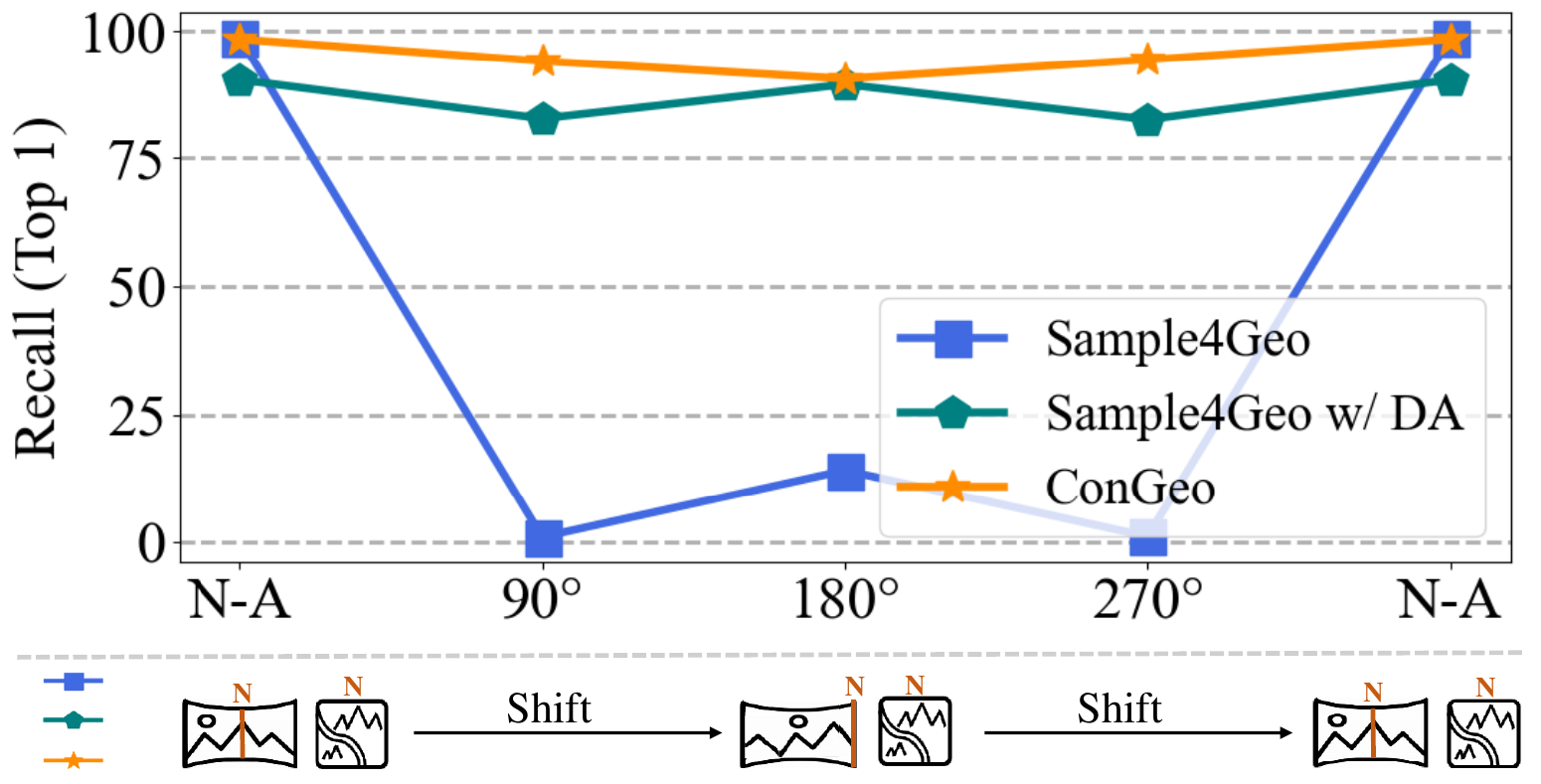}
    \caption{\textbf{ConGeo shows better orientation invariance}. We cyclically shift the ground view with an angle (x-axis) as the model's input to test its retrieval performance. Note that ``N-A'' denotes the North-aligned setting and ``DA'' means data augmentation.}
    \label{fig:asym_sym}
\end{minipage}~
\begin{minipage}[t]{0.49\textwidth}
\centering
\includegraphics[width=0.95\linewidth]{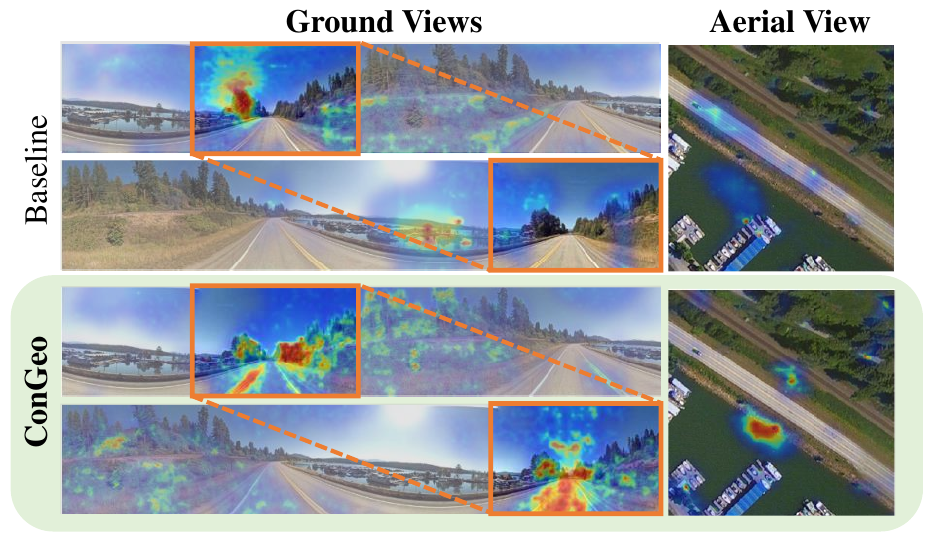}
    \caption{\textbf{ConGeo's activation areas are more consistent across ground view variants}. The Grad-CAM activation maps of the base model and ConGeo on the North-aligned and the unknown orientation setting. The orange box indicates the same area in different ground view variants.}
    \label{fig:attention360}
\end{minipage}

\end{figure}
\begin{figure*}[t]
    \centering
    \includegraphics[width=0.98\linewidth]{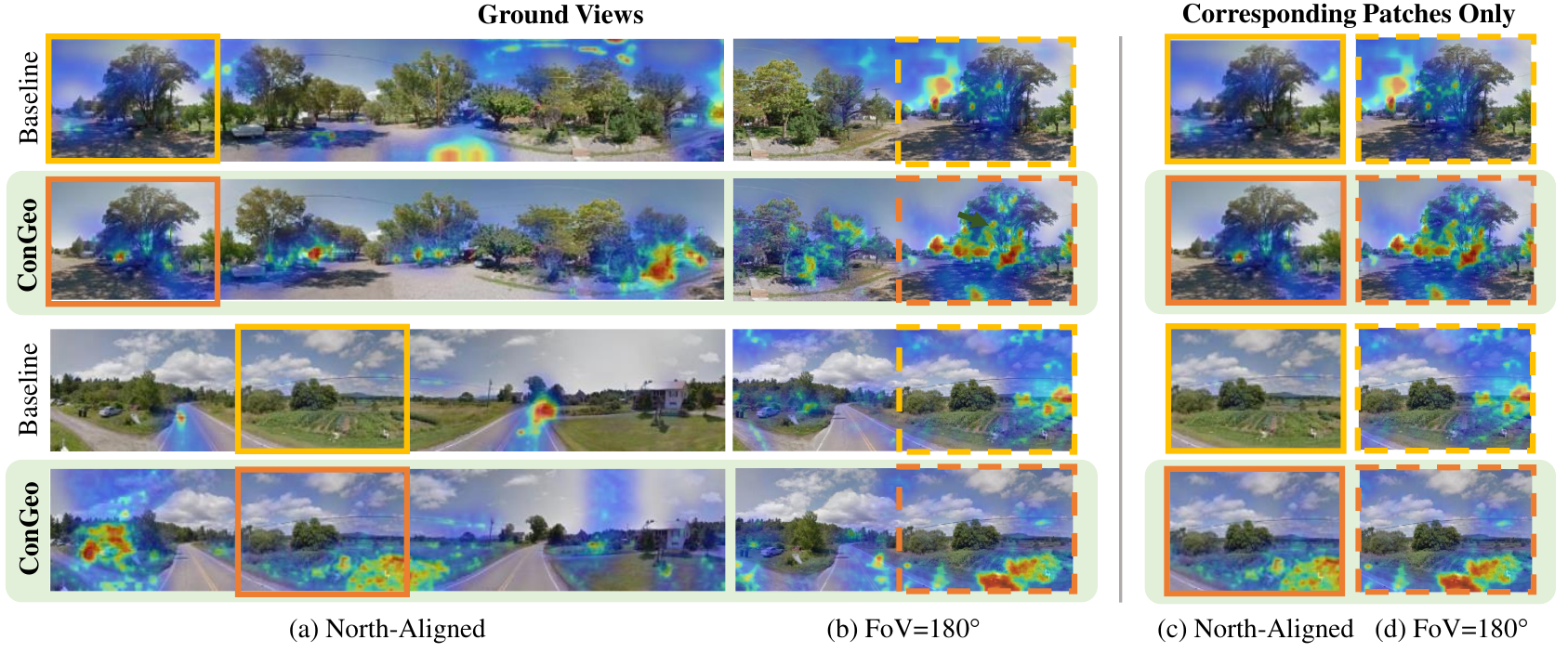}
    \caption{
    \textbf{ConGeo focuses on semantically consistent features for the task and its activated regions are consistent across view variation}. Left: GradCam activations of the base model and ConGeo (green background) on the North-aligned and FoV = 180$^{\circ}$ settings. Right: focus on the regions highlighted by the coloured boxes.
    }
    \label{fig:attention}
\end{figure*}
\section{Analysis: How does ConGeo achieve robustness?}
\label{analysis}

In this section, we analyze the behavior of the base model and ConGeo under ground view variations, to diagnose what led to the base model's collapse and explain the superior performance of ConGeo. We perform orientation invariance analysis to showcase models' vulnerabilities to orientation shifts and visualize activation maps to investigate the models' focus.

\noindent \textbf{Orientation Invariance Analysis.} 
We introduce the concept of orientation invariance to dissect the siamese network's behavior. Let $\Phi$ be a function that maps the input images to predictions, $G:=\{g\}$ be a group of cyclic shift transformations. $I_q$ and $I_r$ are the query and reference input images, respectively. Here, the ground view orientation invariance is defined as:
\begin{equation}
    \Phi\left[g(I_q), I_r\right]=\Phi(I_q, I_r) \quad \forall(I_q, I_r, g) \in(Q, R, G).
\end{equation}
In Fig.~\ref{fig:asym_sym}, we experimentally investigate the orientation invariance of different models by evaluating the retrieval performance when using shifted ground views as input. For the base model trained under North alignment (blue line), the model's recall drops significantly on different orientation angles, indicating a lack of orientation invariance.
Although the model with ground view variants data augmentation (green line) shows an improved orientation invariance compared to the base model with the orientation-specific learning pipeline, the performance of the North-aligned setting drops considerably.
In contrast, ConGeo (orange line) yields consistently high performance both under the cyclically shifted query input and with the North-aligned image pairs. This indicates that the model trained with the proposed contrastive objectives shows better orientation invariance while simultaneously maintaining a strong ability to leverage the spatial correspondence, allowing it to maintain robustness to ground view variations.

\noindent \textbf{Activation Map Visualization.} We analyze the activation map of ConGeo and its baseline in Fig.~\ref{fig:attention360} (North-aligned and unknown orientation settings) and Fig.~\ref{fig:attention} (North-aligned and limited FoV settings). We make two key observations. First, the focus of the base model is vulnerable to orientations and FoV variations, while ConGeo's representation is more \emph{robust across view variations}. In Fig.~\ref{fig:attention360}, the ground view shifts make the base model's attention drift from the roadside to random regions that might not carry geospatial information (\textit{e.g.}, sky), while ConGeo consistently highlights regions (\textit{e.g.}, trees) with similar contents under view variations. Second, the base model focuses more on spatial correspondence cues (\textit{e.g.}, road), while ConGeo focuses more on the \emph{semantically consistent objects} in both views (\textit{e.g.}, trees), as shown in Fig.~\ref{fig:attention}. This further demonstrates that enhancing consistency and mitigating shortcuts of spatial correspondence makes the model more robust to view variations. More examples can be found in supplementary materials.

\section{Limitations}
\label{sec:limitation}
We showed that ConGeo leads to significant improvements when facing arbitrarily oriented ground images and diverse FoVs, except in the North-aligned setting. This performance drop is nearly unavoidable, as ConGeo tends to disrupt the over-reliance on spatial correspondence shortcuts. However, by keeping the original learning objectives of the base model (vanilla loss), ConGeo achieves competitive performance when orientation information is available (Table \ref{tab:standard}), and significantly improves the robustness when it is unknown. Additionally, we focus on ground view orientation and FoV variations, but other variations can be envisaged (\textit{e.g.}, zoom, color intensity, blur). We show ConGeo's robustness to some of these variations unseen during training (Table \ref{tab:unseen}). Finally, besides contrastive learning, other ways of aligning modalities (\textit{e.g.}, redundancy reduction) could also be considered. We discuss this in the supplementary materials.

\section{Conclusion}
Tackling situations where the ground view image orientation is unknown, or the FoV is limited, is crucial for real-world applications of cross-view image geo-localization.
We propose ConGeo, a single- and cross-view contrastive method that enhances a model's robustness to ground view variations by aligning image variations with their original representations. 
Experiments on four datasets and activation map analysis demonstrate that ConGeo consistently boosts the performance of state-of-the-art methods by a large margin when facing diverse challenging settings.
With its adaptability to different base models and versatility to accommodate diverse ground view variations, ConGeo strives to be a step towards widening the applicability of geo-localization methods to the real world.

\section*{Acknowledgments}
We thank the anonymous reviewers for their constructive and thoughtful comments. We thank Haoyuan Li, Zimin Xia, Gencer Sümbül, Silin Gao, Valérie Zermatten, Zeming Chen, Tianqing Fang, Gaston Lenczner, Kuangyi Chen, Robin Zbinden, Giacomo May, Riccardo Ricci, Emanuele Dalsasso, and Sepideh Mamooler for providing helpful feedback on earlier versions of this work. We acknowledge the support from the CSC and EPFL Science Seed Fund and the support in part by the National Natural Science Foundation of China (NSFC) under Grant 62271355. AB gratefully acknowledges the support of the Swiss National Science Foundation (No. 215390), Innosuisse (PFFS-21-29), the EPFL Center for Imaging, Sony Group Corporation, and the Allen Institute for AI.

\appendix
\setcounter{section}{0}
\renewcommand{\thesection}{S\arabic{section}}
\setcounter{table}{0}
\renewcommand{\thetable}{S\arabic{table}}
\setcounter{figure}{0}
\renewcommand{\thefigure}{S\arabic{figure}}

\section*{Supplementray Materials}

The supplementary materials contain the following information:
\begin{itemize}
    \item Dataset details (Section~\ref{ssec:dataset}).
    \item Implementation details (Section~\ref{ssec:implementation}).
    \item Detailed method description (Section~\ref{ssec:method}).
    \item Hyper-parameters analysis: about the training FoV angle, loss weights, and the learning rate (Section~\ref{ssec:parameter}).
    \item Additional experiments: architecture analysis (CNN-based and ViT-based encoders), transfer experiments, polar transformation, different paradigms of aligning modalities (Section~\ref{ssec:exp}).
    \item Supplementary results to the main paper: full tables (Section~\ref{ssec:full}).
    \item Visualization: additional samples and failure case analysis (Section~\ref{ssec:vis}).
    \item Discussions: limitations and future works (Section~\ref{ssec:discuss}).
\end{itemize}

\section{Dataset Details}
\label{ssec:dataset}
\noindent \textbf{CVUSA.} The CVUSA dataset~\cite{cvusa_cvpr_2017} contains 35,532 ground-aerial view pairs for training and 8,884 pairs for evaluation. The satellite images are with the size of 750 $\times$ 750 and street-view images are with the size of 224 $\times$ 1232. Both types of images are North-aligned to ensure that the geographical North is located in the upper center of the satellite image and the center of the street view images.

\noindent \textbf{CVACT.} The CVACT dataset~\cite{lending_2019_CVPR} is split into training, validation, and test sets, where training and validation sets are of the same size as CVUSA while the test set has 92,802 image pairs, which is around 10 times larger than the validation set. The image size is larger than the CVUSA dataset, with $1200 \times 1200$ for satellite images and $832 \times 1664$ for ground views.

\noindent \textbf{VIGOR.} The VIGOR dataset~\cite{vigor_cvpr_2021} contains 90,618 aerial-view images and 105,214 ground-view images from four cities: New York, Seattle, San Francisco, and Chicago. The raw image sizes for aerial view and ground view are 640 $\times$ 640 and 2048 $\times$ 1024, respectively. Different from the one-to-one matching in CVUSA and CVACT datasets, each query image in VIGOR can be paired with multiple reference images and vice versa. There are two subsets in VIGOR datasets: same-area and cross-area. In the paper, we use the cross-area subset to evaluate the model's robustness across locations. In the training set, the cross-area setting contains 44,055 aerial view images with 51,520 ground view images from New York and Seattle. The cross-area test set is sampled from San Francisco and Chicago, with 46,563 aerial view images and 53,694 ground view images.

\noindent \textbf{University-1652.} The University-1652 dataset~\cite{zheng2020university} contains different types of cross-view images, including satellite images, unmanned aerial vehicle (UAV) images, and ground view images. In our experiments, we train and evaluate models based on street-satellite matching to demonstrate the robustness of the model on real-world limited FoV images. For street-to-satellite matching, there are 2,579 street images and 951 satellite images. For satellite-to-street matching, there are 701 satellite images and 2,921 street images. The image size is $512 \times 512$.

\begin{table*}[t]
\setlength{\tabcolsep}{3pt}
\footnotesize
    \centering
    \resizebox{0.99\linewidth}{!}{ 
    \begin{tabular}{c | c | c | c | c | c }
    \toprule
    Method & Metric Learning Loss & Architecture  & Optimizer& Weight-sharing & Zero-padding\\
      \hline
    Sample4Geo-CNN~\cite{sample4geo_2023_iccv} & InfoNCE & CNN & AdamW~\cite{adamw_2017_arxiv} & \Checkmark &\XSolidBrush \\
    Sample4Geo-ViT~\cite{sample4geo_2023_iccv} & InfoNCE & ViT & AdamW~\cite{adamw_2017_arxiv} & \XSolidBrush &  \Checkmark \\
    TransGeo~\cite{transgeo_cvpr_2022} & Soft-triplet & ViT & ASAM~\cite{asam_2021_icml} & \XSolidBrush &  \Checkmark\\
    SAIG-D~\cite{zhu2023simplesaigd} & Soft-triplet & CNN \& Attention & AdamW~\cite{adamw_2017_arxiv} & \XSolidBrush & \Checkmark\\
    \bottomrule
    \end{tabular}}
    \caption{A comparison of the basic settings in different base models.}
    \label{tab:models}
\end{table*}

\section{Implementation Details}
\label{ssec:implementation}
\noindent \textbf{Data Preprocessing.} We followed the data preprocessing methods used in the base models. For example, when we use Sample4Geo~\cite{sample4geo_2023_iccv} as our base model, we resize the ground view images to $140 \times 768$, and the aerial view images to $384 \times 384$ for CVUSA and CVACT. For the VIGOR dataset, we resize the ground view to $384 \times 768$ and the satellite image to $384 \times 384$. As for the University-1652 dataset, all images are resized from $512 \times 512$ to $384 \times 384$. 
Note that for models utilizing TransGeo, SAIG-D, and Sample4Geo[ViT] architectures, we employ zero-padding in the area after FoV cropping for both training and testing inputs, enabling the learnable positional encoding to adapt to inputs with varying FoVs.
Unless specified, we retain the corresponding data augmentation in different base models~\cite{sample4geo_2023_iccv, transgeo_cvpr_2022, zhu2023simplesaigd} for fair comparisons. For those using Sample4Geo as the base model, the data augmentation includes dropout, color jitter, flipping on both ground and aerial views, rotation of satellite images, and the corresponding shift on the ground view images.

\noindent \textbf{Model and Training Details.} 
As described in the paper, for different base models, we follow their default settings respectively, including model architecture, weight-sharing operations, and basic loss \textit{etc.} to ensure a fair comparison. A comparison of the detailed settings in different base models is shown in Table.~\ref{tab:models}.
For instance, for experiments based on Sample4Geo[CNN], the weights of the ground view encoder and the aerial view encoder are shared, while for others (Sample4Geo[ViT] and other base models), the weights are not shared between the two encoders. 
For Sample4Geo[CNN], instead of a batch size of 128 on multiple GPUs, we set the batch size as 16 on a single GPU, which is much smaller. Accordingly, we set the starting learning rate as 0.0001 instead of 0.001 in the default settings. 

\noindent \textbf{Experimental Environment.}
The code was developed using Python version 3.8, PyTorch version 2.0.1, Timm version 0.9.7, and OpenCV-Python version 4.8.1.78. A single NVIDIA GeForce RTX 4090 is utilized for computation.

\section{Detailed Method Description}
\label{ssec:method}
\subsection{Training and Inference}

Here we use the CNN-based Sample4Geo as an example.

\noindent \textbf{Training Phase:} In each mini-batch, we take paired ground-aerial images ($I_Q$, $I_R$) as input, and perform defined transformations $T_{q}$ (\textit{i.e.}, random shift, FoV cropping) to each ground view image $I_q$ ($I_q \in I_Q$), obtaining the transformed ground image $I_q{^*}=T_{q}(I_q | \theta, \alpha)$. We also perform augmentations in the base model to $I_R$ (described in the data preprocessing of Section~\ref{ssec:implementation}), obtaining $I_R{^*}$. Then, the image batchs $I_Q$, $I_Q{^*}$, $I_R$ and $I_R{^*}$ are encoded by two encoders, a query encoder $E_q$ for encoding $I_Q$, $I_Q{^*}$, and a reference encoder $E_r$ for encoding $I_R$, $I_R{^*}$, with output feature embeddings $Q$, $Q{^*}$, $R$, and $R{^*}$ respectively.
The optimization goal contains four targets, two single-view contrastive objectives, one for aligning the feature space between $Q$ and $Q{^*}$ while another for aligning the feature space between $R$ and $R{^*}$, a cross-view contrastive objective aligning the feature space between $R$ and $Q{^*}$, and a vanilla contrastive objective aligning the feature space between $R$ and $Q$. The total loss is a weighted summation of the four loss functions, where the ablation of weights will be studied in the next section.

\noindent \textbf{Testing Phase:} Depending on different evaluation settings, the query image is transformed accordingly. We use the original full-view ground images and the transformed ground view images for the North-aligned setting and the other two challenging settings (unknown orientation and limited FoV settings), respectively. For each query input and a set of reference aerial images, the trained $E_q$ and $E_r$ are used for feature representation, and then the reference images are ranked as the retrieval results based on the cosine similarity between the query feature and reference features.

\subsection{Training on the University-1652 Dataset}

Street images in the University-1652 dataset are ground view images with unknown orientations and limited FoVs. In order to adapt ConGeo to street images without collecting a full-view panorama, we provide an alternative way of building the ground view single-view contrastive objective (Eq. (1) in the paper). Instead of enforcing the representations of ground view variants to be closer to the original ones, we construct the single-view contrastive objective by emphasizing the proximity between street images obtained from the same geographic locations but depicting different perspectives or angles. Since the street-to-satellite retrieval is a many-to-one matching, there is more than one street view image located in the region of satellite images. Therefore, for each street view sample, we randomly select another street image with the same location to construct the single-view contrastive objective for training.

\section{Hyper-parameter Analysis}
\label{ssec:parameter}
\subsection{Training FoV Angle $\alpha$}
In order to achieve robustness across ground view variations, ConGeo applies a set of transformations on ground view images during training. The transformation includes shifting the ground view images with a random orientation angle $\theta$ and cropping the shifted images with a FoV angle $\alpha$. In the main experiments, we empirically set $\alpha$ to 180$^{\circ}$. Here, we provide detailed results by using different training FoV angles. Results are shown in Table~\ref{tab:trainingfov}. In the table, we report models trained with FoV images from 90$^{\circ}$, 180$^{\circ}$, 360$^{\circ}$, and random FoV from 0$^{\circ}$ to 360$^{\circ}$ on the unknown orientation and FoV settings. It can be seen that ConGeo's performance is robust under different training FoVs, which consistently surpasses the baseline with targeted data augmentation by a large margin.   

\begin{table*}[t]
\setlength{\tabcolsep}{3pt}
\footnotesize
    \centering
    \resizebox{0.99\linewidth}{!}{ 
    \begin{tabular}{c | c c c c | c c c c | c c c c | c c c c | c}
    \toprule
    \multirow{2}{*}{Method($\alpha$)} & \multicolumn{4}{c|}{FoV=360$^{\circ}$}  & \multicolumn{4}{c|}{FoV=180$^{\circ}$}  & \multicolumn{4}{c|}{FoV=90$^{\circ}$}  & \multicolumn{4}{c|}{FoV=70$^{\circ}$} & Avg.\\
      & R@1 & R@5 & R@10 & R@1\% & R@1 & R@5 & R@10 & R@1\% & R@1 & R@5 & R@10 & R@1\% & R@1 & R@5 & R@10 & R@1\% & R@1\\
      \hline
     Sample4Geo+DA & 84.1 & 95.0 & 96.9 & 99.4 & 63.6 & 84.6 & 90.1 & 98.2 & 32.2 & 55.1 & 64.5 & 87.4 & 21.5 & 42.0 & 51.6 & 79.9 & 50.4\\
     \hline
     ConGeo(90) & 81.4 & 93.0 & 95.3 & 98.6 & 92.2 & \textbf{98.1} & \textbf{98.9} & \textbf{99.7} & 55.5 & 75.4 & 81.5 & 93.9 & 35.5 & 54.8 & 61.9 & 81.0 & 66.2 \\
     ConGeo(180) & 85.2 & 95.1 & 96.9 & 98.9 & \textbf{92.3} & 97.9 & 98.7 & \textbf{99.7} & \textbf{55.9} & 73.2 & 79.0 & 90.9 & 37.1 & 55.7 & 62.8 & 81.4 & \textbf{67.6}\\
     ConGeo(360) & \textbf{96.6} & \textbf{98.9} & \textbf{99.2} & \textbf{99.7} & 83.8 & 94.2 & 96.1 & 98.8 & 38.2 & 58.2 & 65.2 & 82.0 & 19.5 & 35.9 & 43.8 & 66.5 & 59.5 \\ 
     ConGeo(0-360) & 63.5 & 80.0 & 85.4 & 95.5 & 82.8 & 94.1 & 96.3 & 99.3 & 55.2 & \textbf{77.3} & \textbf{83.0} & \textbf{94.7} & \textbf{46.4} & \textbf{67.1} & \textbf{75.2} & \textbf{91.6} & 62.0\\

    \bottomrule
    \end{tabular}}
    \caption{Analysis of training ConGeo with different FoV angle $\alpha$: 90$^{\circ}$, 180$^{\circ}$, 360$^{\circ}$, and random angles between 0$^{\circ}$-360$^{\circ}$, on the CVUSA dataset. Sample4Geo+DA means training Sample4Geo with targeted data augmentation (random shift and random FoVs 70$^{\circ}$-360$^{\circ}$).}
    \label{tab:trainingfov}
\end{table*}

\begin{table}[t]
    \begin{minipage}[t]{0.48\linewidth}
        \centering
    \resizebox{1.0\columnwidth}{!}{ 
    \begin{tabular}{c c|  c c c | c c c}
    \toprule
    \multicolumn{2}{c|}{Loss weights}  & \multicolumn{3}{c|}{North-aligned}  & \multicolumn{3}{c}{FoV=90$^{\circ}$} \\
    $w_1$ $w_2$ & $w_3$ &  R@1 & R@10 & R@1\% & R@1 & R@10 & R@1\%\\
    \hline
     0.25 & 0.25 & \textbf{98.7} & \textbf{99.7} & \textbf{99.9} & 51.7 & 75.9 & 89.6 \\
     0.25 & 0.5 & 98.5 & \textbf{99.7} & \textbf{99.9} & 54.0 & 78.2 & \textbf{90.9} \\
     0.5 & 0.25 & 98.3 & \textbf{99.7} & \textbf{99.9} & \textbf{55.9} & \textbf{79.0} & \textbf{90.9} \\
     0.5 & 0.5 & 98.4 & \textbf{99.7} & 99.8 & 55.0 & 77.9 & 90.3 \\  
    \bottomrule
    \end{tabular}}
    \caption{Analysis of loss weights of single- and cross-view contrastive objectives on the CVUSA dataset.}
    \label{tab:lossweights}
    \end{minipage}
    \hfill
    \begin{minipage}[t]{0.48\linewidth}
    \centering
    \renewcommand{\arraystretch}{1.1}
    \resizebox{1.0\columnwidth}{!}{ 
    \begin{tabular}{c | c c c | c c c}
    \toprule
    \multirow{2}{*}{LR} & \multicolumn{3}{c|}{North-aligned}  & \multicolumn{3}{c}{FoV=90$^{\circ}$} \\
     & R@1 & R@10 & R@1\% & R@1 & R@10 & R@1\%\\
    \hline
     0.001 & 97.6 & 99.5 & 99.8 & 41.4 & 66.8 & 83.8 \\ 
     0.0001 & \textbf{98.3} & \textbf{99.7} & \textbf{99.9} & \textbf{55.9} & \textbf{79.0} & \textbf{90.9} \\
     0.00001 & 95.6 & 99.4 & 99.8 & 28.8 & 59.2 & 89.4 \\
    \bottomrule
    \end{tabular}}
    \caption{Analysis of the starting learning rate on the CVUSA dataset.}
    \label{tab:lr}
    \end{minipage}
\end{table}

\subsection{Loss Weights $w_1$, $w_2$ and $w_3$}
We investigate the effect of loss weights on the retrieval performance. In Table~\ref{tab:lossweights}, model performance with different loss weights on North-aligned setting and FoV=90$^{\circ}$ are reported. Results suggest that different loss weights may lead to slight performance waves. Specifically, when we gradually increase the $w_1$, $w_2$, and $w_3$, the performance of R@1 under the North-aligned setting will slightly decrease within 1 point, meanwhile, the limited FoV performance will be improved. When the single-view loss weight $w_1$ and $w_2$ value are set higher, there is a significant improvement on FoV=$90^{\circ}$ with around 4 points on R@1.

\subsection{Learning Rate}
We also study the effects of the starting learning rate on training results. In Table~\ref{tab:lr}, we take the CVUSA dataset for example, and verify the training performance under 0.001, 0.0001, and 0.00001, respectively. Results show that the best performance is obtained under a learning rate of 0.0001. Moreover, we can observe that the performance under the North-aligned setting is robust to learning rates while the limited FoV setting is sensitive to learning rates.

\section{Supplementary Experiments}
\label{ssec:exp}
\subsection{Architecture Analysis}
\label{sec:architecture}
In the main paper, we demonstrate that ConGeo, as a learning objective, can be plugged into different base models, including the CNN-based model~\cite{sample4geo_2023_iccv} and ViT-based model~\cite{transgeo_cvpr_2022}. To better compare those two mainstream architectures for cross-view geo-localization, we use Sample4Geo as the base model and switch the backbone between CNN and ViT in Table~\ref{tab:strucrtures} to test the corresponding performance on the North-aligned settings and FoV=90$^{\circ}$. In other words, the only difference between these two base models is the encoder architecture: Sample4Geo-CNN is with ConvNeXt~\cite{convnext_cvpr_2022} and Sample4Geo-ViT is with Swin Transformer~\cite{liu2021swin}.
From the results, we can see that ConGeo brings considerable improvement to both architectures, between them, the improvement with the CNN-based backbone is larger. We also compare the effect of training FoV angle ($\alpha$) on different architectures. 
The results indicate that CNN-based architecture yields the best overall performance when setting $\alpha$ to 180$^{\circ}$. The ViT-based architecture, although shows less competitive performance compared to the CNN-based one, seems to be more robust with different training angles $\alpha$.
Here, we present preliminary findings regarding the analysis of different architectures under unknown orientations and limited FoV settings. A more comprehensive investigation into this direction could serve as an intriguing topic for future research.

\begin{table}[t]
    \begin{minipage}[t]{0.48\linewidth}
    \setlength{\tabcolsep}{3pt}
    \centering
    \resizebox{0.98\columnwidth}{!}{ 
    \begin{tabular}{c | c c c | c c c}
    \toprule
    \multirow{2}{*}{Method} & \multicolumn{3}{c|}{North-aligned}  & \multicolumn{3}{c}{FoV=90$^{\circ}$} \\
     & R@1 & R@10 & R@1\% & R@1 & R@10 & R@1\%\\
    \hline
     Sample4Geo-CNN & \textbf{98.7} & \textbf{99.8} & \textbf{99.9} & 2.5 & 9.8 & 26.7 \\ 
     ConGeo-CNN[180] & 98.3 & 99.7 & \textbf{99.9} & \textbf{55.9} & \textbf{79.0} & \textbf{90.9} \\
     ConGeo-CNN[Random] & 98.1 & 99.6 & 99.8 & 31.0 & 58.0 & 77.2 \\
     \hline
     Sample4Geo-ViT & \textbf{97.6} & \textbf{99.6} & \textbf{99.9} & 1.7 & 6.5 & 20.8 \\ 
     ConGeo-ViT[180] & 96.5 & 99.3 & 99.8 & \textbf{51.0} & \textbf{77.8} & \textbf{91.9} \\
    ConGeo-ViT[Random] & 96.9 & 99.3 & 99.8 & 48.4 & 74.9 & 90.7 \\ 
    \bottomrule
    \end{tabular}}
    \caption{Architecture analysis. The comparison of the base model and ConGeo (training FoV angle $\alpha$ set to 180 and random, respectively) with different backbones: CNN-based and ViT-based.}
    \label{tab:strucrtures}
    \end{minipage}
    \hfill
    \begin{minipage}[t]{0.48\linewidth}
    \setlength{\tabcolsep}{3pt}
    \centering
    \renewcommand{\arraystretch}{1.3}
    \resizebox{0.98\columnwidth}{!}{ 
    \begin{tabular}{c | c c c | c c c}
    \toprule
    \multirow{2}{*}{Method} & \multicolumn{3}{c|}{North-aligned}  & \multicolumn{3}{c}{FoV=90$^{\circ}$} \\
     & R@1 & R@10 & R@1\% & R@1 & R@10 & R@1\%\\
    \hline
     Sample4Geo w/o Polar & \textbf{98.7} & \textbf{99.8} & \textbf{99.9} & 2.5 & 9.8 & 26.7 \\ 
     \textbf{ConGeo} w/o Polar & 98.3 & 99.7 & \textbf{99.9} & \textbf{55.9} & \textbf{79.0} & \textbf{90.9} \\
     \hline
     Sample4Geo w/ Polar & \textbf{98.8} & \textbf{99.7} & \textbf{99.8} & 3.9 & 12.0 & 27.4 \\ 
     \textbf{ConGeo} w/ Polar & 98.4 & 99.6 & \textbf{99.8} & \textbf{39.0} & \textbf{67.7} & \textbf{86.6} \\
    \bottomrule
    \end{tabular}}
    \caption{Comparison between ConGeo and the base model when the aerial images are with or without polar transformation.}
    \label{tab:polar}
    \end{minipage}
\end{table}
\begin{table*}[t]
\setlength{\tabcolsep}{3pt}
\footnotesize
    \centering
    \resizebox{0.99\linewidth}{!}{ 
    \begin{tabular}{c | c c c c | c c c c | c c c c | c c c c}
    \toprule
    \multirow{3}{*}{Method} & \multicolumn{8}{c|}{CVUSA$\rightarrow$CVACT} & \multicolumn{8}{c}{CVACT$\rightarrow$CVUSA} \\
    \cline{2-17}
    &  \multicolumn{4}{c|}{FoV=360$^{\circ}$}  & \multicolumn{4}{c|}{FoV=90$^{\circ}$}  & \multicolumn{4}{c|}{FoV=360$^{\circ}$}  & \multicolumn{4}{c}{FoV=90$^{\circ}$} \\
      & R@1 & R@5 & R@10 & R@1\% & R@1 & R@5 & R@10 & R@1\% & R@1 & R@5 & R@10 & R@1\% & R@1 & R@5 & R@10 & R@1\%\\
    \hline
     Sample4Geo & 3.4 & 6.4 & 8.1 & 17.6 & 0.2 & 0.8 & 1.5 & 7.6 & 3.1 & 5.7 & 7.0 & 14.5 & 0.3 & 0.9 & 1.7 & 7.7 \\ 
     \textbf{ConGeo} & \textbf{4.2}& \textbf{9.8} & \textbf{13.3} & \textbf{33.6} & \textbf{1.0} & \textbf{3.4} & \textbf{5.2} & \textbf{19.8} & \textbf{5.7} & \textbf{12.2} & \textbf{17.1} & \textbf{39.6} & \textbf{2.2} & \textbf{5.3} & \textbf{8.1} & \textbf{24.9} \\
    \bottomrule
    \end{tabular}}
    \caption{Transfer results between CVUSA dataset and CVACT dataset on the unknown orientation setting and limited FoV setting.}
    \label{tab:transfer}
\end{table*}

\subsection{Polar Transformation Analysis}
As we reviewed in Section~\ref{sec:related_works} in the main paper, polar transformation is commonly used to improve the models' performance on the North-aligned setting~\cite{jointloc_2020_cvpr, transgeo_cvpr_2022}. Polar transformation transfers the aerial images to the polar coordinates and narrows the view gap between aerial images and ground images. We compare the performance between the base model and ConGeo with the original satellite images or with the polar transformed satellite images in Table~\ref{tab:polar}. Results indicate ConGeo can consistently improve the robustness, whether the aerial images are polar-transformed or not. Note that polar transformation enhances the spatial correspondences in the model. Therefore, compared with ConGeo without polar transformation, ConGeo with polar transformation is more likely to maintain strong performance in the North-aligned setting, while being less competitive in the limited FoV settings.

\subsection{Transfer between CVUSA and CVACT Datasets}
The transfer analysis between CVUSA and CVACT datasets is often reported to show the generalization ability of the model on different datasets. We also report the transfer results of the base model and ConGeo on the unknown orientation setting and limited FoV setting in Table~\ref{tab:transfer}. Results suggest ConGeo can significantly improve the robustness of the base model (R@1\% gains on average 20.6\% on FoV=360$^{\circ}$ over the two settings).

\subsection{Different Paradigms of Aligning Modalities}
ConGeo requires the alignment of different modalities. Besides contrastive learning (CL), other methods, e.g., direct feature alignment (FA), and redundancy reduction (RR), can also be used to align modalities. To investigate this, we also compare different paradigms of alignment modalities using InfoNCE~\cite{oord2018representation} (CL), Barlow Twins~\cite{zbontar2021barlow} (RR) and Consine similarity (FA) to represent each paradigm respectively. The results shown in Table~\ref{tab:align} suggest that aligning modalities helps to improve robustness, but due to the huge cross-modal view gap and single-view information asymmetry, FA fails to reduce the feature distance, RR struggles to find the shared features, while CL excels in learning by comparing.

\begin{table}[t]
\setlength{\tabcolsep}{5pt}
\footnotesize
    \centering
    \resizebox{0.6\columnwidth}{!}{ 
    \begin{tabular}{c| c| c c | c c }
    \toprule
    \multirow{2}{*}{Paradigm} & \multirow{2}{*}{Method} & \multicolumn{2}{c|}{North-Aligned}  & \multicolumn{2}{c}{FoV=90$^{\circ}$}\\
     & & R@1 & R@1\% & R@1 & R@1\% \\
    \hline
     FA & Cosine Similarity &  86.6 & 99.6 & 31.2 & 81.9\\
     RR & Barlow Twins~\cite{zbontar2021barlow} &  85.1 & 99.2 & 41.4 & 82.1 \\
     CL [\textbf{ConGeo}] & InfoNCE~\cite{oord2018representation}  &  \textbf{98.3} & \textbf{99.9} & \textbf{55.9} & \textbf{90.9}\\ 
    \bottomrule
    \end{tabular}}  
    \caption{Comparison of different paradigms of aligning modalities. ``FA'', ``RR'' and ``CL'' denote direct alignment, redundancy reduction and contrastive learning, respectively.}
    \label{tab:align}
\end{table}

\section{Supplementary Results}
\label{ssec:full}

\subsection{Full Table of the North-aligned Setting}
We provide the full comparison of the North-aligned setting on CVUSA and CVACT datasets in Table~\ref{tabs:standard}. Among all the methods, ConGeo achieves competitive performance on the North-aligned setting on both datasets meanwhile maintaining robustness across different challenging settings. The results indicate the effectiveness and versatility of the proposed contrastive objectives.

\begin{table*}[t]
\setlength{\tabcolsep}{7pt}
\small
   \begin{center}
    \resizebox{0.99\textwidth}{!}{ 
    \begin{tabular}{l|cccc|cccc|cccc}
       \toprule
       \multirow{2}{*}{Method} & \multicolumn{4}{c|}{CVUSA} & \multicolumn{4}{c|}{CVACT Val}& \multicolumn{4}{c}{CVACT Test}  \\ 
        & R@1 & R@5 & R@10  & R@1\% &  R@1 & R@5 & R@10  & R@1\%&  R@1 & R@5 & R@10  & R@1\%\\ \hline
       CVM-Net\cite{hu2018cvmCVMNET} & 22.47 & 49.98 & 63.18  & 93.62 & 82.49 & 92.44 & 93.99 & 97.32 & - & - & - & -\\
       LPN~\cite{wang2021eachlpn} & 85.79 & 95.38 & 96.98 & 99.41 & 79.99 & 90.63 & 92.56 & - & - & - & - & - \\
       SAFA~\cite{safa_nips_2019} & 89.84 & 96.93 & 98.14 & 99.64 & 81.03 & 92.80 & 94.84 & - & - & - & - & -\\
       CVFT \cite{shi2020optimalCVFT} & 61.43 & 84.69 & 90.49 & 99.02 & 61.05 & 81.33 & 86.52 & 95.93 & - & - & - & - \\
       DSM \cite{jointloc_2020_cvpr}  & 91.96 & 97.50 & 98.54 & 99.67 & 82.49 & 92.44 & 93.99 & 97.32 & - & - & - & - \\
       CDE \cite{toker2021comingCDE} & 92.56 & 97.55 & 98.33 & 99.57 & 83.28 & 93.57 & 95.42 & 98.22 & 61.29 & 85.13 & 89.14 & 98.32 \\
       L2LTR \cite{yang2021crossl2ltr} &94.05 & 98.27 & 98.99 & 99.67 & 84.89 & 94.59 & 95.96 & 98.37 & 60.72 & 85.85 & 89.88 & 96.12 \\
       SEH \cite{guo2022softSEH} & 95.04 & 98.31 & 98.92 & 99.76 & 85.13 & 93.84 & 95.24 & 97.97 & - & - & - & -\\
       TransGeo~\cite{transgeo_cvpr_2022} & 94.08 & 98.36 & 99.04 & 99.77 & 84.95 & 94.14 & 95.78 & 98.37 & - & - & - & -\\
       GeoDTR~\cite{zhang2023crossGeoDTR} & 95.43 & 98.86 & 99.34 & 99.86 & 86.21 & 95.44 & 96.72 &  98.77 & 64.52 & 88.59 & 91.96 & 98.74\\ 
       SAIG-D~\cite{zhu2023simplesaigd} & 96.34 & 99.10 & 99.50 & 99.86 & 89.06 & 96.11 & 97.08 & 98.89 & 67.49 & 89.39 & 92.30 & 96.80\\ 
       Sample4Geo~\cite{sample4geo_2023_iccv} & 98.68 & 99.68 & 99.78 & 99.87 & 90.81 & 96.74 & 97.48 & 98.77  & 71.51 & 92.42 & 94.45 & 98.70\\
       \hline
     \textbf{ConGeo} & 98.27 & 99.59 & 99.70 & 99.86 & 90.12 & 95.69 & 96.56 & 98.24 & 71.67 & 91.61 & 93.50 & 98.30\\
       \bottomrule
    \end{tabular}
    }
   \end{center}
     \caption{The full table of comparison of ConGeo and state-of-the-art methods on the North-aligned setting on the CVUSA and CVACT.}
    \label{tabs:standard}
\end{table*}

\begin{table*}[t]
\setlength{\tabcolsep}{3pt}
\footnotesize
    \centering
    \resizebox{\linewidth}{!}{ 
    \begin{tabular}{c | c c| c c | c c c c | c c c c | c c c c | c c c c}
    \toprule
    \multirow{2}{*}{$\mathcal{L}_{\text{single-r}}$} &\multicolumn{2}{c|}{$\mathcal{L}_{\text{single-q}}$} & \multicolumn{2}{c|}{$\mathcal{L}_{\text{cross}}$}  & \multicolumn{4}{c|}{FoV=360$^{\circ}$} & \multicolumn{4}{c|}{FoV=180$^{\circ}$} & \multicolumn{4}{c|}{FoV=90$^{\circ}$}  & \multicolumn{4}{c}{FoV=70$^{\circ}$} \\
     & Shift & FoV & Shift & FoV & R@1 & R@5 & R@10 & R@1\%  & R@1 & R@5 & R@10 & R@1\% & R@1 &  R@5 & R@10 & R@1\% & R@1 & R@5 & R@10 & R@1\% \\
    \hline
      & &   & & & 16.3 & 26.1 & 31.4 & 51.7 & 4.1 & 8.4 & 11.3 & 30.4 & 2.5 & 6.7 & 9.8 & 26.7 & 1.5 & 4.6 & 6.7 & 20.4 \\
      \checkmark &  & & & & 30.1 & 48.4 & 57.1 & 79.1 & 15.7 & 28.2 & 34.7 & 60.0 & 7.7 & 16.2 & 21.5 & 43.3 & 5.8 & 12.9 & 17.4 & 38.9 \\ 
      \checkmark & \checkmark & & & & 89.7 & 96.9 & 97.9 & 99.5 & 44.1 & 70.3 & 78.9 & 93.4 & 17.8 & 35.8 & 44.8 & 73.0 & 12.0 & 26.5 & 33.7 & 56.3\\ 
     \checkmark & \checkmark & \checkmark & &  & 28.6 & 45.9 & 54.8 & 83.3 & 37.9 & 56.1 & 62.4 & 81.2 & 20.5 & 34.9 & 40.8 & 59.9 & 14.2 & 26.3 & 32.3 & 53.4 \\
      \checkmark & & & \checkmark & \checkmark & 73.8 & 88.6 & 91.9 & 97.3 & 91.5 & 97.8 & \textbf{98.8} & \textbf{99.7} & 40.2 & 69.6 & 75.7 & 89.0 & 29.6 & 48.3 & 55.9 & 76.0\\ 
     \checkmark & \checkmark &  & \checkmark & & \textbf{96.5} & \textbf{98.9} & \textbf{99.4} & \textbf{99.7} & 81.7 & 93.4 & 95.7 & 98.8 & 35.8 & 56.2 & 63.5 & 81.6 & 20.3 & 37.0 & 44.3 & 67.6 \\
     \checkmark & \checkmark & \checkmark & \checkmark & \checkmark & 85.2 & 95.1 & 96.9 & 98.9 & \textbf{92.3} & \textbf{97.9} & 98.7 & \textbf{99.7} &  \textbf{55.9} & \textbf{73.2} & \textbf{79.0} & \textbf{90.9} & \textbf{37.1} & \textbf{55.7}  & \textbf{62.8} & \textbf{81.4} \\
    \bottomrule
    \end{tabular}
    }   
    \caption{The full table of ablation studies on FoV=70$^{\circ}$, 90$^{\circ}$, 180$^{\circ}$, and 360$^{\circ}$ on the CVUSA dataset. ``Shift'' and ``FoV'' mean cyclic shift and FoV cropping. ``Single'' and ``Cross'' denote single-view contrastive objective cross-view contrastive objective, respectively.}
    \label{tabs:ablation1}
\end{table*}
\begin{table*}[!t]
\setlength{\tabcolsep}{3pt}
\footnotesize
    \centering
    \resizebox{\linewidth}{!}{ 
    \begin{tabular}{c c c| c c c c | c c c c | c c c c | c c c c}
    \toprule
    \multicolumn{3}{c|}{Augmentation} & \multicolumn{4}{c|}{FoV=360$^{\circ}$}  & \multicolumn{4}{c|}{FoV=180$^{\circ}$}  & \multicolumn{4}{c|}{FoV=90$^{\circ}$}  & \multicolumn{4}{c}{FoV=70$^{\circ}$} \\
    Shift & FoV & Rotate & R@1 & R@5 & R@10 & R@1\% & R@1 & R@5 & R@10 & R@1\% & R@1 & R@5 & R@10 & R@1\% & R@1 & R@5 & R@10 & R@1\%\\
    \hline
      & & & 16.3 & 26.1 & 31.4 & 51.7 & 4.1 & 8.4 & 11.3 & 30.4 & 2.5 & 6.7 & 9.8 & 26.7 & 1.5 & 4.6 & 6.7 & 20.4 \\
     & \checkmark & & 9.6 & 17.6 & 22.5 & 46.8 & 6.0 & 10.8 & 13.7 & 29.8 & 3.4 & 6.8 & 9.5 & 24.7 & 2.6 & 5.6 & 7.1 & 21.1\\ 
     \checkmark & & & \textbf{93.1} & \textbf{97.6} & \textbf{98.2} & \underline{99.1} & \underline{73.8} & \underline{88.0} & 90.9 & 95.8 & 35.1 & 54.2 & 61.2 & 77.6 & 18.9 & 34.0 & 41.5 & 62.5 \\ 
     \checkmark & \checkmark & & 84.1 & 95.0 & 96.9 & \textbf{99.4} & 63.6 & 84.6 & 90.1 & \underline{98.2} & 32.2 & 55.1 & 64.5 & \underline{87.4}  & 21.5 & 42.0 & 51.6 & \underline{79.9}\\ 
    \checkmark & \checkmark & \checkmark& \underline{89.0} & \underline{96.1} & \underline{97.3} & 98.9 & 71.8 & 87.9 & \underline{91.3} & 97.2 & \underline{39.3} & \underline{59.9} & \underline{67.9} & 85.7 & \underline{29.8} & \underline{50.0} & \underline{58.0} & 79.3 \\ 
         \hline
     \multicolumn{3}{c|}{ConGeo} & 85.2 & 95.1 & 96.9 & 98.9 & \textbf{92.3} & \textbf{97.9} & \textbf{98.7} & \textbf{99.7} &  \textbf{55.9} & \textbf{73.2} & \textbf{79.0} & \textbf{90.9} & \textbf{37.1} & \textbf{55.7}  & \textbf{62.8} & \textbf{81.4} \\
    \bottomrule
    \end{tabular}}
    \caption{The full table of comparison with different data augmentation methods on FoV=70$^{\circ}$, 90$^{\circ}$, 180$^{\circ}$, and 360$^{\circ}$ on the CVUSA dataset. ``Shift'' applies random cyclical shift, ``FoV'' means applying random FoV cropping (from 70$^{\circ}$ to 360$^{\circ}$) to ground images as data augmentation for training, `Rotate'' randomly rotating aerial images with an angle in \{90$^{\circ}$, 180$^{\circ}$, 270$^{\circ}$\} as data augmentation.}
    \label{tab:ablation2}
\end{table*}

\subsection{Full Tables of Ablation Studies}
We provide the results of component ablations and the comparison with different data augmentation methods in Table~\ref{tabs:ablation1} and Table~\ref{tab:ablation2}, respectively. 

As shown in Table~\ref{tabs:ablation1}, through the full results provided, we can further confirm the individual effectiveness of each component in the proposed ConGeo. Moreover, the combination of single-view and cross-view contrastive objectives can further boost the model performance indicating that they can complement each other's alignment. Besides, we can see from the results that merely applying ``Shift'' transformations (FoV=$360^{\circ}$) to the ground view images in single-view and cross-view contrastive objectives achieves the best performance, showing that the model performance on specific FoVs can be further improved by using the evaluation FoV for training like in previous methods.

Experimental results in Table~\ref{tab:ablation2} confirm the superiority of contrastive objectives over data augmentation methods. Although data augmentation methods can improve the model performance in specific settings, they struggle to obtain robustness across various settings. For example, when randomly shifting the panorama image as data augmentation, the retrieval performance under FoV=$360^{\circ}$ can be marginally better than ConGeo, while it trails ConGeo by about 20 points at other settings.

\section{Visualization Results}
\label{ssec:vis}
\subsection{Retrieval Results}
\begin{figure*}[t]
    \centering
    \includegraphics[width=0.98\linewidth]{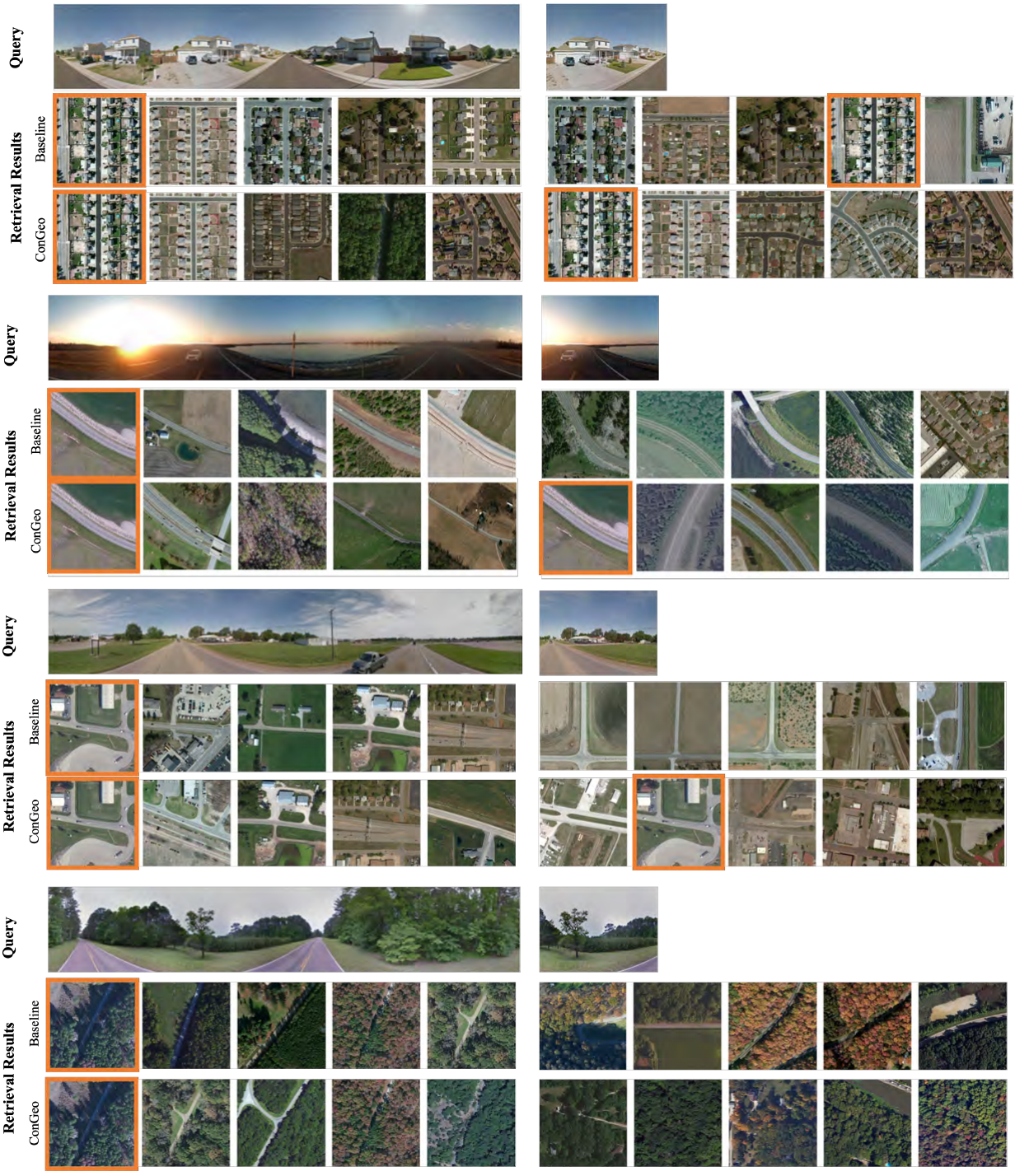}
    \caption{Retrieval results of the baseline method and ConGeo in the North-aligned setting and FoV=$90^{\circ}$. Retrieval results are ranked by the similarity score. Images marked in yellow denote the correct retrieval result.}
    \label{sfig:results}
\end{figure*}

More visualization results of the North-aligned setting and FoV=$90^\circ$ on the CVUSA dataset are provided in Fig.~\ref{sfig:results}. We make two main observations: 

1) Compared to ConGeo, the top candidates retrieved by the baseline method are of similar geometric structure (\textit{e.g.} road direction).
When provided with full-view panoramas, both the baseline and the proposed ConGeo can find the target aerial images. When using the limited FoV setting, the spatial correspondence between the ground view images and the paired aerial images is partially broken, which brings difficulties to the models. As shown in the second and third examples, the baseline model's results are fairly limited to similar road structures (North-south direction), however, with the contrastive objective, the top retrieval candidates exhibit more diverse structures (arbitrarily orientated roads).

2) Compared to the baseline, the top candidates retrieved by ConGeo are of distinguishable similar content. It indicates that, besides the geometric information, ConGeo also focuses on the consistent semantic content shared by both views. This is particularly noticeable when the FoV is limited. For example, in the first example, ConGeo notices the buildings in the ground views are with the same roof color and returns results that are consistent with that. However, even given color clues in the query image, the baselines' retrieval results are with different roof colors. The second and third examples show that ConGeo can retrieve images with the distinguishable semantic cues in images (\textit{e.g.}, water and building), while the baseline fails to capture them.

For better understanding, we plot the rank of ground truth reference images in the retrieval results of the base model and of ConGeo on the left part of Fig.~\ref{sfig:distribution}, when FoV=$90^\circ$. For example, $rank=0$ means the reference image is correctly retrieved by the corresponding model. Compared with the baseline model, the distribution of ConGeo is gathered on the head and is much sparser in the tail, which demonstrates that ConGeo not only achieves better performance among metrics but also improves the feature representation of hard examples.

\subsection{Visualization of Activation Maps}

\begin{figure*}[t]
    \centering
    \includegraphics[width=0.99\linewidth]{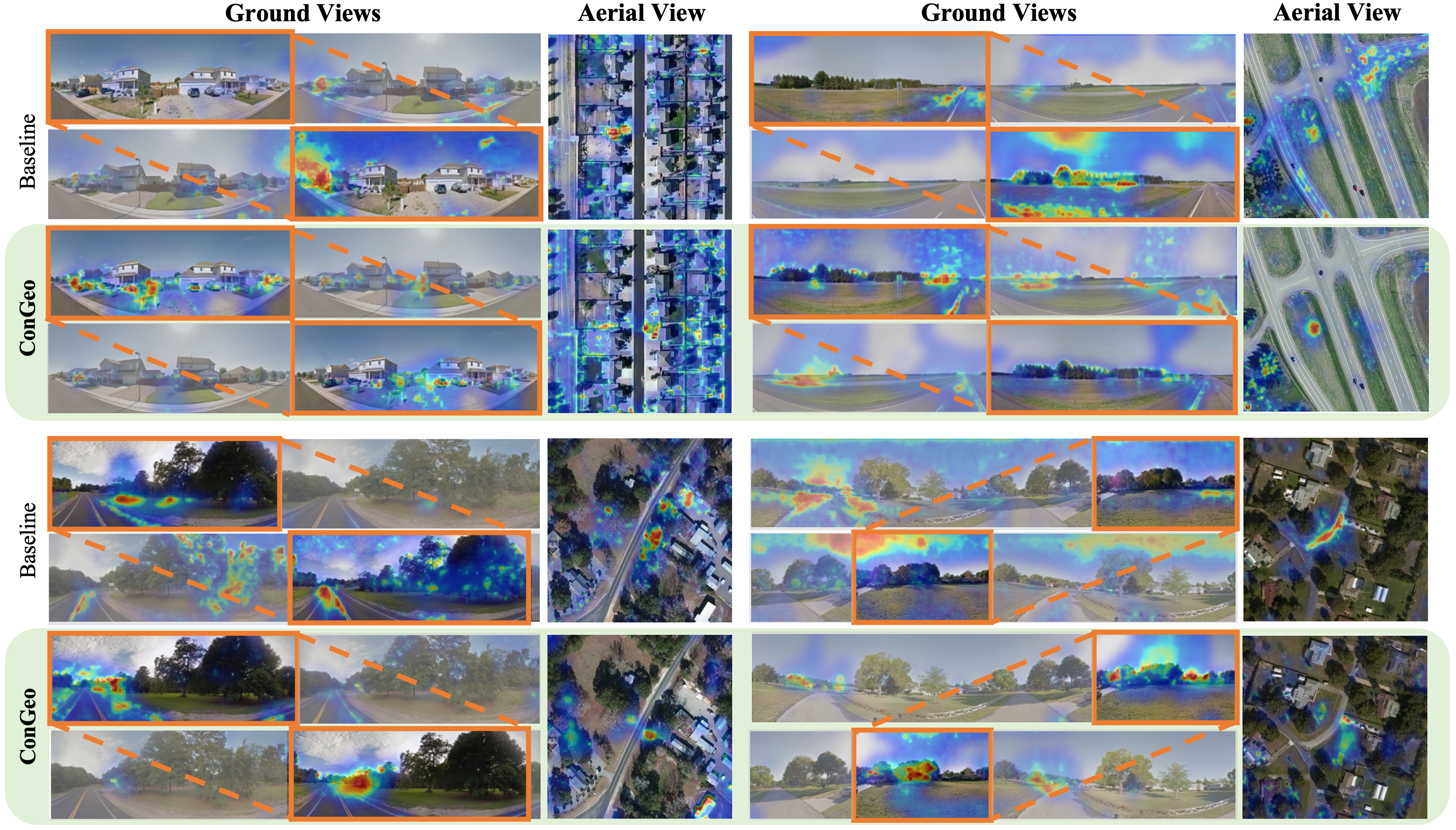}
    \caption{Additional examples of the Grad-CAM activation maps of the base model and ConGeo on the North-aligned setting (the first row for each sample) and the unknown orientation setting (the second row for each sample). The orange box indicates the same area in different ground view variants.}
    \label{sfig:consistent}
\end{figure*}

\begin{figure*}[t]
    \centering
    \includegraphics[width=0.98\linewidth]{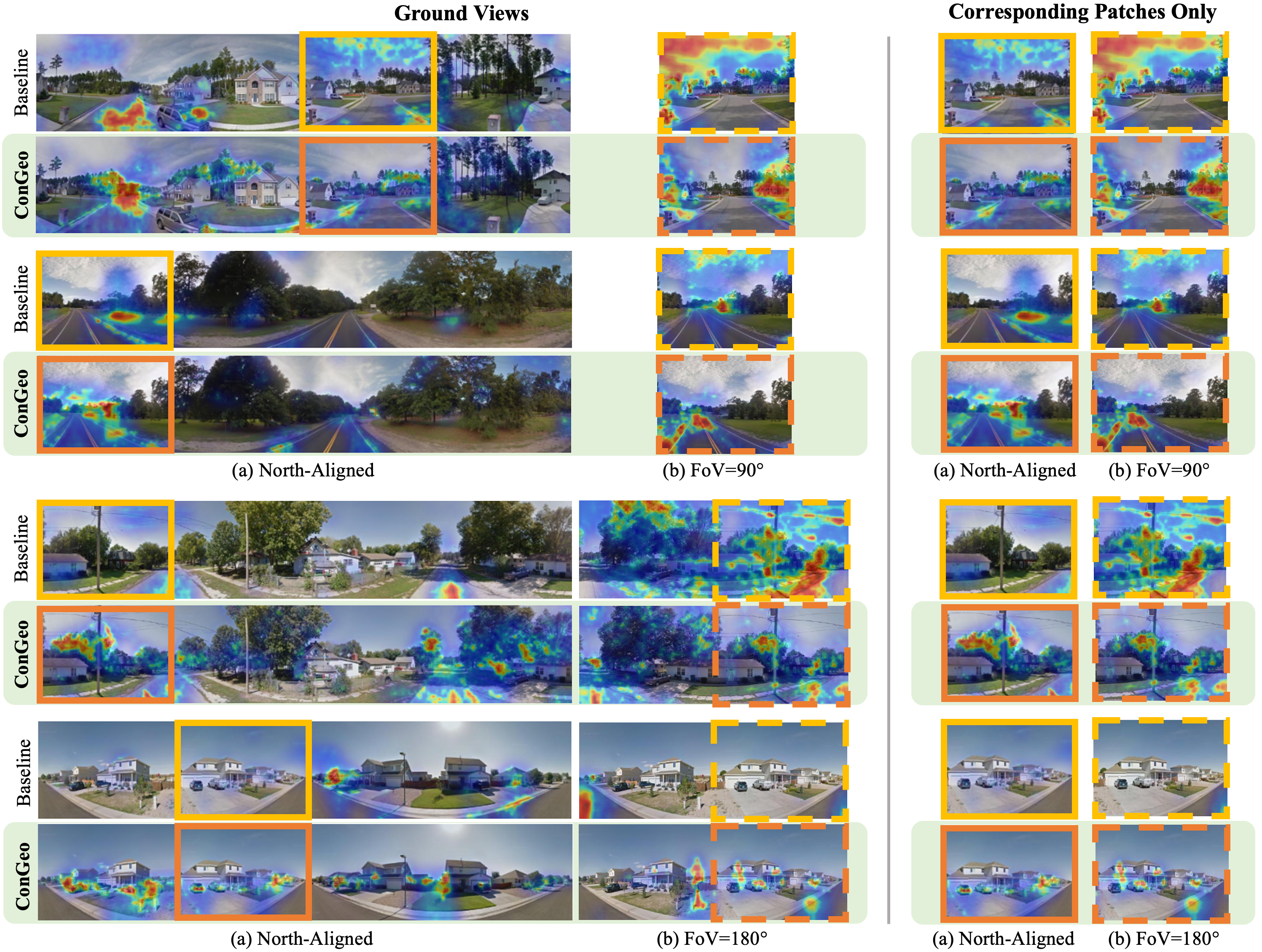}
    \caption{Additional examples of the Grad-CAM activation maps of the base model (top row) and ConGeo (bottom row) on the North-aligned setting (a) and limited FoV setting (b) on the left. Corresponding patches from the two settings are shown on the right.}
    \label{sfig:activation}
\end{figure*}

We provide more cases of activation map visualization in Fig.~\ref{sfig:consistent} and Fig.~\ref{sfig:activation}. As mentioned in the paper, we use Grad-CAM \cite{jacobgilpytorchcam} to visualize input regions that contribute the most to the model's predictions, on the CVUSA dataset. The feature maps are extracted from the ConvNeXt blocks of the model and superimposed with the images. 

In Fig.~\ref{sfig:consistent}, we compare the activation maps of different orientations of the panorama. Since the shifted image shares same content with its prototype, the activation areas of a robust model are supposed to be similar across different orientations. We find that the base model's focus is vulnerable to orientations, while ConGeo's attention is more robust. For example, in the first example (the left of the first row), ConGeo's activation areas are distributed in similar regions of the buildings, while the baseline fails to keep consistent across the North-aligned and unknown orientation settings. In the last example (the right of the last row), the ground view variations make the base model's attention drift to the sky which might not carry geospatial information, while ConGeo consistently highlights the road trees.

In Fig.~\ref{sfig:activation}, there are comparisons between the baseline and the proposed ConGeo on the North-aligned setting and the limited FoV setting. Compared with the baseline, the activation areas of ConGeo focus more on the coherent semantic objects across views. For example, in the first example (the first two rows), the activation areas of ConGeo (the second row) are mostly on the building along the road while for the baseline model (the first row), the attention is misleading (on the sky). This problem will be severe when FoV=$90^\circ$, the activation areas of the baseline model are distributed on the sky.
In addition, we observe that ConGeo alleviates the model's over-reliance on spatial correspondence. For example, from the third example, we can see that the baseline model mostly relies on the road structure to match the full-view panorama with the aerial view image. When the spatial correspondence is broken as in the FoV images, the model fails to find the cues merely based on the learned geometric structure. However, for ConGeo, with the help of semantic content, the model learns to retrieve the aerial image with the incomplete geometric structure. The activation map analysis further confirms that ConGeo is more robust to view variations.

\subsection{Failure Cases: What Kind of Samples Are More Sensitive?}
\label{sec:fail}
Our method may fail when very limited information is provided in the query image. The first case is that the query image is of a very small FoV. As can be seen from Table~\ref{tab: fov} of the main paper, the retrieval performance under very limited FoV (\textit{e.g.} 70$^{\circ}$) is significantly lower than other settings. The second case is that the query image contains merely single and common layouts. Some samples exhibit greater sensitivity to the robustness test, making it more challenging for the model to uphold its robustness on such samples. The model will struggle to distinguish the corresponding aerial image from similar counterparts given common layouts. 

\begin{figure*}[t]
    \centering
    \includegraphics[width=0.95\linewidth]{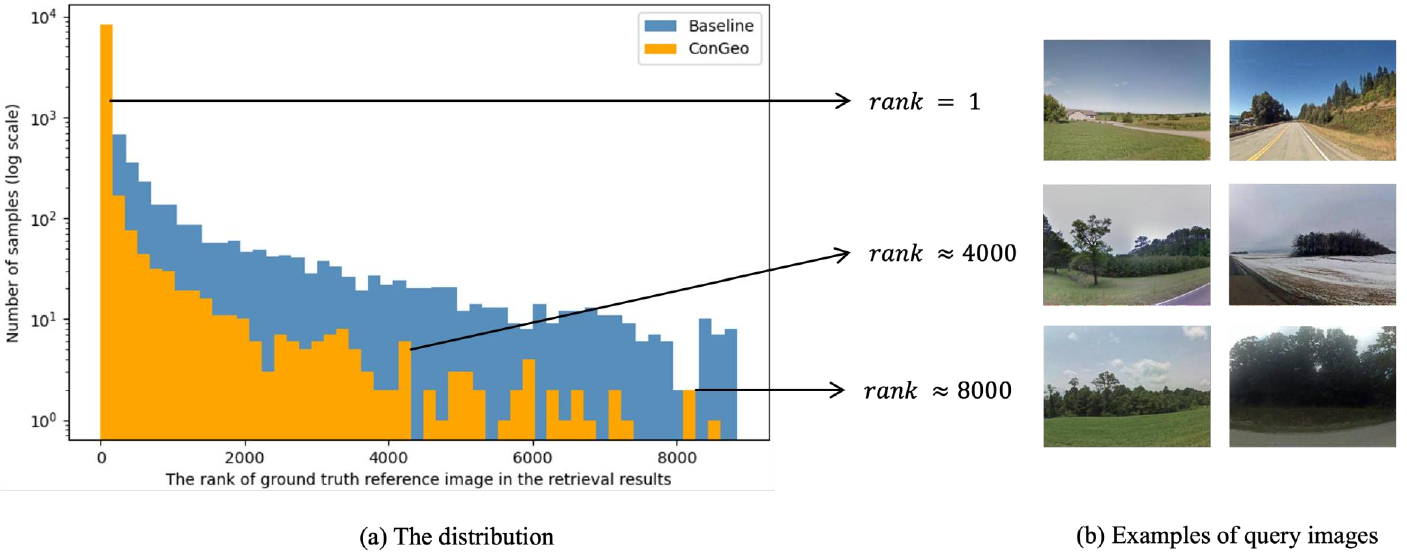}
    \caption{Distribution of the rank of ground truth reference images in the retrieval results when FoV=$90^\circ$ (left) and some examples of query images according to the retrieval results of ConGeo (right).}
    \label{sfig:distribution}
\end{figure*}

We show the examples on the right part of Fig.~\ref{sfig:distribution} when the rank of ground truth annotations are 14,000, and 8,000. In general, when the query image only contains grass or trees, the similarity score of the ground truth aerial image may be drowned in other similar candidates. Analysing the sensitivity of different samples can also help improve model robustness and can be an interesting direction for future research.

\section{Discussion}
\label{ssec:discuss}
In this work, we propose ConGeo to boost the base models' robustness across ground view variations. 
Without specialized training, ConGeo outperforms state-of-the-art methods on unknown orientation and limited FoV benchmarks and demonstrates adaptability to three different base models and generalizability to various unseen ground view variations. 
However, as mentioned in Section~\ref{sec:limitation} of the main paper, several aspects and challenges remain to be addressed for future research.

First, compared to models trained specifically for the North-aligned setting, ConGeo faces an almost unavoidable slight performance drop in this setting. Indeed, when performing orientation-specific training, North-alignment is a key assumption and models often take shortcuts by using the spatial correspondence in the data stemming from this alignment. However, ConGeo aims to relax it and achieve robustness in more challenging settings, where such shortcuts will reduce the model's generality. Moreover, it is also worth noting that the North-aligned setting is a special case of FoV=360$^{\circ}$, where the orientation angle is set to 0$^{\circ}$. Therefore, the performance drop in the North-aligned setting is almost unavoidable when improving the unknown orientation setting, which also indicates a trade-off between orientation-specific or FoV-specific spatial correspondence and the robust semantic information across ground view variations. Despite this slight decrement, ConGeo serves as a practical solution to real-world scenarios: When the orientation is known, ConGeo performs comparably to the state-of-the-art methods, while the model significantly improves the performance when the orientation is unknown.

Second, in this work, we increase the models' robustness by proposing a new learning pipeline; however, other aspects of the model can face robustness issues and could be discussed, such as network architecture and sample-level information. As mentioned in Section~\ref{sec:plugin} of the main paper and Section~\ref{sec:architecture} in this supplementary material, ConGeo can consistently improve the models' robustness regardless of the backbones. Future research can further study the differences between different backbones and different components in the base models that lead the model to prefer learning shortcuts in the data over robust features. As discussed in the previous work in image classification~\cite{hendrycks2019benchmarking}, some backbones are more likely to rely on spurious correlations or cues inessential to the object. Moreover, as we showed and discussed in Section~\ref{sec:fail} in this supplementary material, some samples with less salient semantic information might be more sensitive to ground view variations. This is further supported by prior research on content-based adversarial attacks~\cite{chen2024content}, which indicates that certain elements within an image are more susceptible to variations. Thus, further research might improve robustness across ground view variations from the data perspective.

Finally, in this work, we mainly focus on ground view orientation and reduced FoV variations. However, we also validate the effectiveness of the model under several unseen variations (Table~\ref{tab:unseen} in the main paper), such as blur and zooming. Expanding the robust geo-localization across a broader range of ground view variations (\textit{e.g.}, nighttime or extreme weather conditions) can further widen the applicability of this research to real-world use cases.

\newpage
\newpage

\bibliographystyle{splncs04}
\bibliography{reference}

\end{document}